%% file: _main.tex
\pdfoutput=1
\documentclass[10pt,twocolumn,letterpaper]{article}

\usepackage[pagenumbers]{cvpr} %

\usepackage[accsupp]{axessibility}
\usepackage[utf8]{inputenc} %
\usepackage[T1]{fontenc}    %
\usepackage{url}            %
\usepackage{booktabs}       %
\usepackage{amsfonts}       %
\usepackage{nicefrac}       %
\usepackage{microtype}      %
\usepackage{mathtools}
\usepackage[pdftex]{graphicx}
\usepackage{amsmath}
\usepackage{amssymb}
\usepackage{amsthm}
\usepackage{colortbl}
\usepackage{wrapfig}
\usepackage{bm}
\usepackage[binary-units]{siunitx}
\usepackage[dvipsnames]{xcolor}
\usepackage{array}
\usepackage{enumitem}
\usepackage{multirow}
\usepackage{makecell}
\usepackage{algorithm}
\usepackage{algpseudocode}
\usepackage{xspace}
\usepackage{rotating}
\usepackage[colorlinks,breaklinks=true,pagebackref]{hyperref}
\hypersetup{
    pdfauthor={Jérôme Rony, Jean-Christophe Pesquet, Ismail {Ben Ayed}},
    pdftitle=Proximal Splitting Adversarial Attack for Semantic Segmentation,
    linkcolor=BrickRed,
    citecolor=Green
}

\definecolor{Gray}{gray}{0.9}
\renewcommand{\algorithmiccomment}[1]{\textcolor{gray}{\bgroup\hfill//~#1\egroup}}
\input{notations}

\newtheorem{proposition}{Proposition}
\theoremstyle{definition}

\newtheorem*{definition*}{Definition}
\theoremstyle{remark}
\newtheorem*{remark}{Remark}

\def\cvprPaperID{2877} %
\def\confName{CVPR}
\def\confYear{2023}

\begin{document}

\title{\vspace{-1.75em}Proximal Splitting Adversarial Attack for Semantic Segmentation}

\author{%
  Jérôme Rony\\
  ÉTS Montréal\\
  \and
  Jean-Christophe Pesquet \\
  Centre de Vision Numérique \\
  Université Paris-Saclay, CentraleSupélec, Inria\\
  \and
  Ismail {Ben Ayed} \\
  ÉTS Montréal\\
  \and
  {\small Code: \url{https://github.com/jeromerony/alma_prox_segmentation}}
}

\maketitle

\input{0_abstract}
\input{1_intro}
\input{2_related_works}
\input{3_preliminaries}

\input{4_method}
\input{5_experiments}
\input{6_results}
\input{7_conclusion}

\clearpage
{
\small
\bibliographystyle{ieee_fullname}
\bibliography{biblio}
}

\clearpage
\input{9_appendix}

\end{document}

%% file: notations.tex
\DeclareMathAlphabet{\mathalphm}{OMS}{cmsy}{m}{n}
\DeclareMathAlphabet{\mathalphb}{OMS}{cmsy}{b}{n}

\newcommand{\linf}{\ell_\infty}
\newcommand{\norm}[1]{\left\lVert#1\right\rVert}

\newcommand{\pertset}{\Lambda}

\newcommand{\x}{\bm{x}}
\newcommand{\xadv}{\tilde{\x}}
\newcommand{\pert}{\bm{\delta}}
\newcommand{\ytrue}{\bm{y}}
\newcommand{\ytarget}{\bm{t}}
\newcommand{\g}{\bm{g}}
\newcommand{\p}{\bm{p}}
\newcommand{\proj}[1]{\mathcal{P}_{#1}}
\newcommand{\prox}[1]{\mathrm{prox}_{#1}}

\newcommand{\y}{\bm{y}}
\newcommand{\z}{\bm{z}}

\newcommand{\Y}{\mathcal{Y}}
\newcommand{\X}{\mathcal{X}}

\newcommand{\vrho}{\bm{\rho}}
\newcommand{\vmu}{\bm{\mu}}
\newcommand{\vlambda}{\bm{\lambda}}
\newcommand{\vm}{\bm{m}}
\newcommand{\vd}{\bm{d}}
\newcommand{\vy}{\bm{y}}
\newcommand{\vv}{\bm{v}}
\newcommand{\metric}{\mathsf{H}}

\def\vx{{\bm{x}}}

\makeatletter
\DeclareRobustCommand\onedot{\futurelet\@let@token\@onedot}
\def\@onedot{\ifx\@let@token.\else.\null\fi\xspace}

\def\eg{\emph{e.g}\onedot} 
\def\ie{\emph{i.e}\onedot}

\def\wrt{w.r.t\onedot} 
\def\etal{\emph{et al}\onedot}
\makeatother

\newcommand{\R}{\mathbb{R}}

\DeclareMathOperator*{\argmin}{arg\,min}
\DeclareMathOperator*{\argmax}{arg\,max}

\DeclareMathOperator*{\st}{s.t.}

\newcommand{\sL}{{\mathcal{L}}}

\makeatletter

\newcommand\Autoref[1]{\@first@ref#1,@}
\def\@throw@dot#1.#2@{#1}%
\def\@set@refname#1{%
    \edef\@tmp{\getrefbykeydefault{#1}{anchor}{}}%
    \xdef\@tmp{\expandafter\@throw@dot\@tmp.@}%
    \ltx@IfUndefined{\@tmp autorefnameplural}%
         {\def\@refname{\@nameuse{\@tmp autorefname}s}}%
         {\def\@refname{\@nameuse{\@tmp autorefnameplural}}}%
}
\def\@first@ref#1,#2{%
  \ifx#2@\autoref{#1}\let\@nextref\@gobble%
  \else%
    \@set@refname{#1}%
    \@refname~\ref{#1}%
    \let\@nextref\@next@ref%
  \fi%
  \@nextref#2%
}
\def\@next@ref#1,#2{%
   \ifx#2@ and~\ref{#1}\let\@nextref\@gobble%
   \else, \ref{#1}%
   \fi%
   \@nextref#2%
}

\makeatother

%% file: 0_abstract.tex
\begin{abstract}
Classification has been the focal point of research on adversarial attacks, but only a few works investigate methods suited to denser prediction tasks, such as semantic segmentation.
The methods proposed in these works do not accurately solve the adversarial segmentation problem and, therefore, overestimate the size of the perturbations required to fool models.
Here, we propose a white-box attack for these models based on a proximal splitting to produce adversarial perturbations with much smaller $\ell_\infty$ norms.
Our attack can handle large numbers of constraints within a nonconvex minimization framework via an Augmented Lagrangian approach, coupled with adaptive constraint scaling and masking strategies.
We demonstrate that our attack significantly outperforms previously proposed ones, as well as classification attacks that we adapted for segmentation, providing a first comprehensive benchmark for this dense task.
\end{abstract}

%% file: 1_intro.tex
\section{Introduction}

\begin{figure*}
    \centering
    \begin{subfigure}{0.404\linewidth}
        \includegraphics[width=\linewidth]{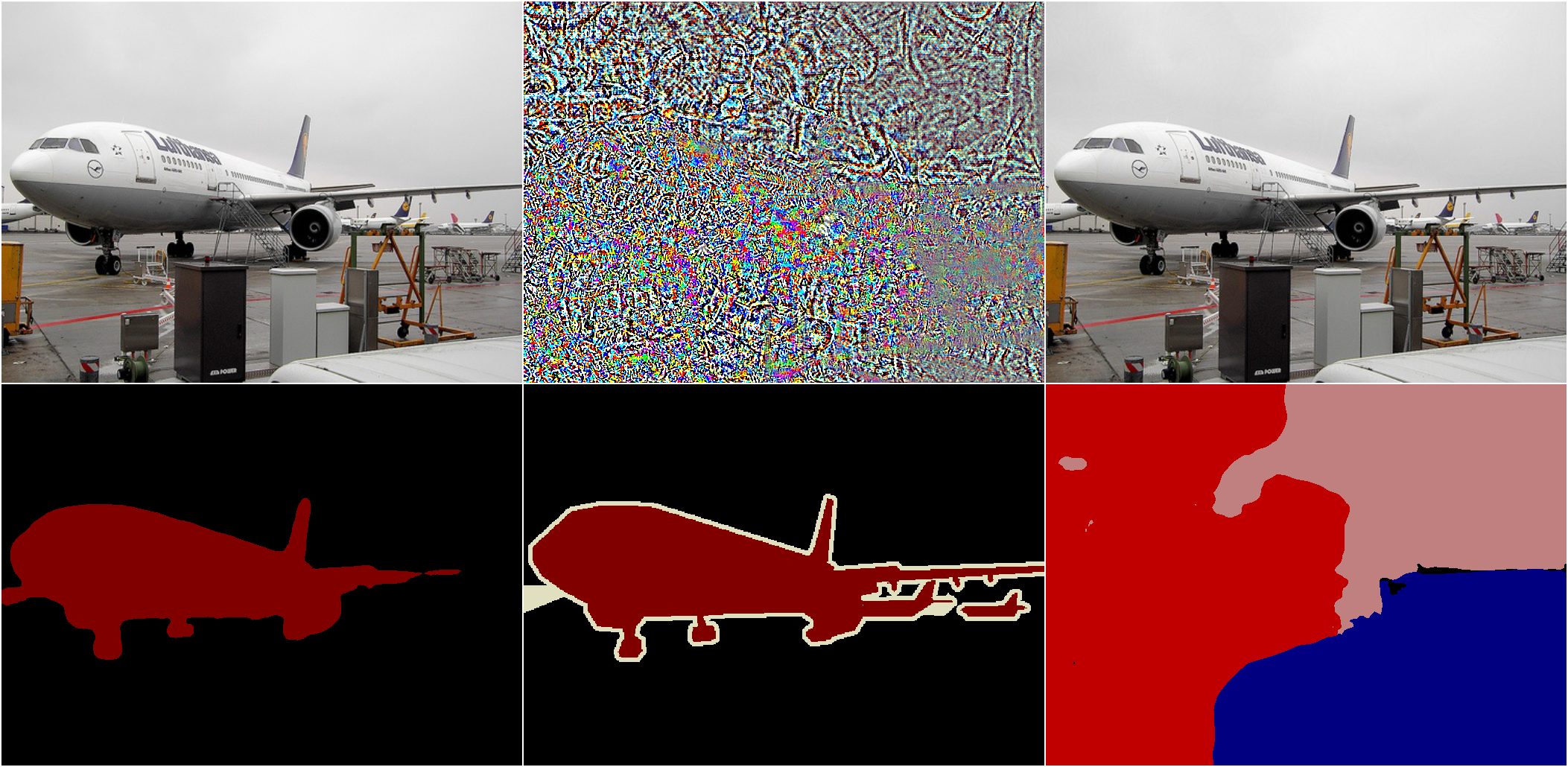}
        \caption{Pascal VOC 2012: $\norm{\pert}_\infty=\nicefrac{0.67}{255}$}
    \end{subfigure}
    \hfill
    \begin{subfigure}{0.591\linewidth}
        \includegraphics[width=\linewidth]{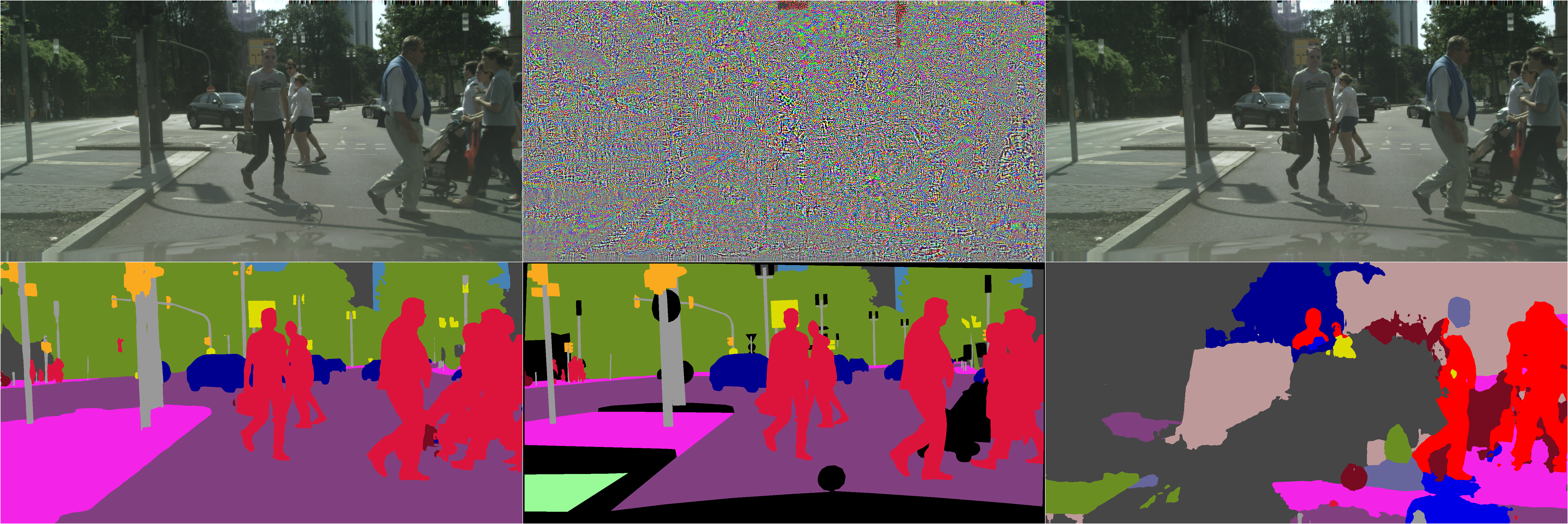}
        \caption{Cityscapes: $\norm{\pert}_\infty=\nicefrac{0.53}{255}$}
    \end{subfigure}
    \caption{Untargeted adversarial examples for FCN HRNetV2 W48 on Pascal VOC 2012 and Cityscapes. In both cases, more than $99\%$ of pixels are incorrectly classified. For each dataset, left is the original image and its predicted segmentation, middle is the amplified perturbation and the ground truth segmentation and right is the adversarial image with its predicted segmentation. For Pascal VOC 2012, the predicted classes are \textit{TV monitor} (blue), \textit{person} (beige) and \textit{chair} (bright red).}
    \label{fig:untargeted_seg}
\end{figure*}

Research on white-box adversarial attacks has mostly focused on classification tasks, with several methods proposed over the years for $\ell_p$-norms \cite{kurakin2017adversarial,moosavi2016deepfool,dong2018boosting,carlini2017towards,rony2019decoupling,yao2019trust,brendel2019accurate,croce2020minimally,croce2020reliable,rony2021augmented,matyasko2021pdpgd}. In contrast, the literature on adversarial attacks for segmentation tasks has been much scarcer, with few works proposing attacks \cite{xie2017adversarial, cisse2017houdini, ozbulak2019impact}. The lack of studies on adversarial attacks in segmentation may appear surprising because of the prominence of this computer vision task in many applications where a semantic understanding of image contents is needed. In many safety-critical application areas, it is thus crucial to assess the robustness of the employed segmentation models. 

Although segmentation is treated as a per-pixel classification problem, designing adversarial attacks for semantic segmentation is much more challenging for several reasons. First, from an optimization perspective, the problem of adversarial example generation is more difficult. In a classification task, producing minimal adversarial examples is a nonconvex constrained problem with a single constraint. In a segmentation task, this optimization problem now has multiple constraints, since at least one constraint must be addressed for each pixel in the image. For a dataset such as Cityscapes \cite{cordts2016cityscapes}, the images have a size of $2\,048{\times}1\,024$, resulting in more than 2 million constraints. Consequently, most attacks originally designed for classification cannot be directly extended to segmentation. For instance, penalty methods such as C\&W \cite{carlini2017towards} cannot tackle multiple constraints since they rely on a binary search of the penalty weight.

Second, the computational and memory cost of generating adversarial examples can be prohibitive. White-box adversarial attacks usually rely on computing the gradient of a loss \wrt the input. In segmentation tasks, the dense outputs result in high memory usage to perform a backward propagation of the gradient. For reference, computing the gradients of the logits \wrt the input requires ${\sim}\SI{22}{\gibi\byte}$ of memory for FCN HRNetV2 W48 \cite{wang2020deep} on Cityscapes with a $2\,048{\times}1\,024$ image. Additionally, most recent classification adversarial attacks require between 100 and 1\,000 iterations, resulting in a run-time of up to a few seconds per image \cite{rony2019decoupling, pintor2021fast} (on GPU) depending on the dataset and model. For segmentation, this increases to tens or even hundreds of seconds per image with larger models.

In this article, we propose an adversarial attack to produce minimal adversarial perturbations \wrt the $\linf$-norm, for deep semantic segmentation models. Building on Augmented Lagrangian principles, we introduce adaptive strategies to handle a large number of constraints (\ie ${>}10^6$). Furthermore, we tackle the nonsmooth $\linf$-norm minimization with a proximal splitting instead of gradient descent. In particular, we show that we can efficiently compute the proximity operator of the sum of the $\linf$-norm and the indicator function of the space of possible perturbations. This results in an adversarial attack that significantly outperforms the DAG attacks \cite{xie2017adversarial}, in addition to several classification attacks that we were able to adapt for segmentation. We propose a methodology to evaluate adversarial attacks in segmentation, and compare the different approaches on the Cityscapes \cite{cordts2016cityscapes} and Pascal VOC 2012 \cite{pascal-voc-2012} datasets with a diverse set of architectures, including well-known DeepLabV3+ models \cite{chen2018encoder}, the recently proposed transformer-based SegFormer \cite{xie2021segformer}, and the robust DeepLabV3 DDC-AT model from \cite{xu2021dynamic}. The proposed approach yields outstanding performances for all models and datasets considered in our experiments. For instance, our attack finds untargeted adversarial perturbations with $\linf$-norms lower than $\nicefrac{1}{255}$ on average for all models on Cityscapes. With this attacker, we provide better means to assess the robustness of deep segmentation models, which has often been overestimated until now, as the existing attacks could only find adversarial examples with larger norms.

%% file: 2_related_works.tex
\section{Related Works}
\label{sec:related_works}

While the literature on minimal adversarial attacks for classification is vast, the research on attacks for segmentation is much less developed. The main work on adversarial attacks for segmentation is done by Xie \etal \cite{xie2017adversarial}. It proposes a simple algorithm to generate adversarial perturbations for dense prediction tasks, including object detection and segmentation, called the Dense Adversary Generation (DAG) attack. In this attack, the rescaled gradient of the loss \wrt the input is added to the current perturbation, until the stopping criterion is reached, \ie a given percentage of pixels is adversarial. In each iteration, the total loss is the sum of the losses over pixels that are not adversarial. This can be seen as a form of greedy algorithm. See \autoref{sec:dag_attack} for the complete algorithm of the DAG attack and a discussion on the stopping criterion used. In practice, this attack is quite efficient, however, it simply accumulates gradients until the stopping criterion is reached. Therefore, it does not minimize the norm considered. Cisse \etal propose the Houdini attack for several tasks \cite{cisse2017houdini}, including segmentation. The goal of this approach is to maximize a surrogate loss for a given perturbation budget (\ie constraint on the $\linf$-norm), hence not producing minimal perturbations. More recently, Ozbulak \etal studied adversarial examples on a medical image segmentation task \cite{ozbulak2019impact}. They propose a targeted attack to minimize the $\ell_2$-norm, which is a regular penalty method. The weight of the penalty terms is fixed to 1, however, leading to large perturbations.

Other works study the robustness of segmentation models against adversarial attacks \cite{fischer2017adversarial, arnab2018robustness, kang2020adversarial}. In these works, the authors use FGSM \cite{kurakin2017adversarial} or an iterative version of FGSM. However, FGSM is not a minimization attack and is known to provide rough robustness evaluations. This leads to largely overestimated robustness results on both Pascal VOC 2012 and Cityscapes in \cite{arnab2018robustness}.

Even though most adversarial attacks were designed for classification, some may be adapted for segmentation tasks. In particular, $\linf$ attacks that do not rely on projections onto an estimated decision boundary can be used for segmentation (as opposed to DeepFool \cite{moosavi2016deepfool} or FAB \cite{croce2020minimally}). These attacks are PGD \cite{madry2018towards}, FMN \cite{pintor2021fast} and PDPGD \cite{matyasko2021pdpgd}. Note that PDPGD \cite{matyasko2021pdpgd} relies on a proximal splitting method, but uses the AdaProx algorithm \cite{melchior2019proximal}; the latter is appealing but, unlike the prox-Newton algorithm it is inspired from \cite{becker2012quasi}, introduces a mismatch between the scaling in the computation of the proximity operator and the step-size of the gradient step. 
The convergence study of such an algorithm would be quite challenging in the non-convex case. It is also known that, even in the convex case, when such a mismatched algorithm converges, the asymptotic point differs from the solution to the original optimization problem \cite{savanier2022unmatched}.

%% file: 3_preliminaries.tex
\section{Preliminaries}

Let $\x\in\X$ be an input image with its corresponding label map $\ytrue\in\Y$. Usually, in computer vision problems, $\X=[0, 1]^{C\times H\times W}$ and $\Y = \{1, \dots, K\}^{H\times W}$, where $H$ and $W$ are the height and width of the image, $C$ is the number of channels (\eg 3 for RGB images), and $K$ is the number of labelling classes. Our objective is to fool the model $f\colon \X \rightarrow \R^{K\times H\times W}$ producing \textit{logits} $\z=f(\x) = \big(f(\x)_{k,i,j}\big)_{k,i,j}\in\R^{K\times H\times W}$. This means that
we are looking for a perturbation vector $\pert$ such that, for every pixel
$(i,j)\in\{1, \dots, H\}\times\{1,\dots,W\}$, the maximum value of
$(f(\x+\pert)_{k,i,j})_{1\le k \le K}$ is reached for a label different from the true one $\ytrue_{i,j}$. In addition, the perturbation should 
satisfy the value range constraints: $\x+\pert \in \X$. We denote $\pertset$ the set of admissible perturbations, so here $\pertset=[0, 1]^{C\times H\times W}-\x$. 
For simplicity, with a slight abuse of notation, we index the pixels with $i\in \{1, \dots, d\}$ where $d=HW$ is the total number of pixels. 

\paragraph{Problem formulation}
The minimal adversarial perturbation problem for the $\linf$-norm can be formulated as follows:
\begin{equation}
\begin{aligned}
\label{eq:minimal_adv_optim}
    \min_{\pert}\quad & \norm{\pert}_\infty \quad \\
    \st\quad &\argmax_{k\in \{1,\ldots,K\}} f(\x + \pert)_{k,i} \neq \bm{y}_i, \quad i=1,\dots,d, \\
    & \x + \pert \in \X.
\end{aligned}
\end{equation}
The $d$ $\argmax$ constraints correspond to the misclassification of each pixel. The $\x + \pert \in \X$ constraint corresponds to producing a perturbation that results in a valid image, and can be re-written as $\pert\in\pertset$. Note that this constraint can also be encoded in the objective using the indicator function:
\begin{equation}
    \iota_\pertset(\pert) = \begin{cases}0 \quad &\text{if} \quad \pert\in\pertset; \\ {+}\infty \quad &\text{else,}\end{cases}
\end{equation}
resulting in the minimization of $\norm{\pert}_\infty + \iota_\pertset(\pert)$.

\paragraph{Equivalent problem}
As is common in several attacks \cite{carlini2017towards,croce2020reliable,rony2021augmented}, we replace the non-differentiable misclassification constraints by differentiable ones, to make it compatible with first-order optimization methods. Here, the $\argmax$ constraints are replaced by constraints on the Difference of Logits Ratio (DLR) \cite{croce2020minimally}, or rather the $\mathrm{DLR^+}$ \cite{rony2021augmented}:
\begin{equation}
\begin{aligned}
\label{eq:minimal_adv_optim_dlr}
    \min_{\pert}\quad & \norm{\pert}_\infty\\
    \st\quad &\mathrm{DLR^+}(f(\x+\pert)_i, \ytrue_i) + \varepsilon \leq 0, \ i=1,\dots,d, \\
    & \pert \in \pertset,
\end{aligned}
\end{equation}
where $\mathrm{DLR^+}(\z,y)=\frac{\z_y - \max\limits_{i\neq y}\z_i}{\z_{\pi_1}-\z_{\pi_3}}$, with $\pi_i$ the index of the $i$-th largest logit and $\varepsilon$ a small positive constant.

\paragraph{Augmented Lagrangian method} One way to handle the misclassification constraints is to use an Augmented Lagrangian approach \cite{rony2021augmented}, which in the classification setting, performs a gradient descent on the following quantity:
\begin{equation}
\label{eq:alma_loss}
    D(\x+\pert, \x) + P\Big(\mathrm{DLR^+}(g(\x+\pert)), \rho, \mu\Big),
\end{equation}
where $D$ is a discrepancy measure (\eg $\ell_2$-norm, LPIPS \cite{zhang2018unreasonable}), $P$ is a penalty-Lagrangian function parametrized by a parameter $\rho$ and a multiplier $\mu$, and $g:\X\rightarrow\R$ is a classification model. In \cite{rony2021augmented}, the penalty multiplier $\mu$ is updated after every gradient descent iteration to increase or decrease the weight of penalty, and eventually satisfy the misclassification constraint.

%% file: 4_method.tex
\section{Proposed Method}
\label{sec:proposed_method}

Our segmentation attack is built on the general Augmented Lagrangian principle, which has led to competitive performances in the ALMA classification attack \cite{rony2021augmented}.
It provides an efficient solution to challenging nonconvex problems arising when designing an attacker, and we will see that it can be extended to cope with multiple constraints.
This can be achieved by introducing one penalty per constraint, with its associated parameter $\rho$ and multiplier $\mu$. 
Denoting $\vd = \mathrm{DLR^+}(f(\x+\pert), \ytrue) \in \R^d$, we tackle problem \eqref{eq:minimal_adv_optim} by minimizing the following objective:
\begin{equation}
\label{eq:alma_prox_loss}
    \underbrace{\vphantom{\big(}\norm{\pert}_\infty + \iota_\pertset(\pert)}_{h_1} + \underbrace{\vphantom{\big(}\bm{m}^\top P(\vd,\vrho,\vmu)}_{h_2},
\end{equation}
where $\vm\in\{0,1\}^d$ is a binary mask (detailed in the following section), $(\vrho,\vmu)\in\R_{++}^d\times\R_{++}^d$ are the penalty parameters and multipliers associated with each constraint, and $P$ is the penalty function applied componentwise. This formulation raises many technical challenges in the context of segmentation, when dealing with millions of constraints.
Additionally, gradient based optimization does not accommodate nonsmooth functions such as the $\linf$-norm.
In this work, we bring several modifications to make it (\textit{i}) suitable for segmentation and (\textit{ii}) applicable to the $\linf$-norm.

Our attack, called ALMA $\mathrm{prox}$, consists in minimizing \eqref{eq:alma_prox_loss} using a proximal splitting \cite{combettes2011proximal} to handle the nonsmooth term $h_1$ and an Augmented Lagrangian method to satisfy the constraints by minimizing $h_2$ using a gradient descent.
In \autoref{sec:constraint_strategies}, we introduce the adaptive constraint strategies to handle large numbers of constraints in the Augmented Lagrangian framework, and in \autoref{sec:proximal_splitting}, we detail the proximal splitting iteration.
The complete algorithm of our attack is provided in \autoref{sec:alma_prox}.

\subsection{Adaptive constraints strategies}
\label{sec:constraint_strategies}

\paragraph{Constraint masking} 

In segmentation tasks, some regions are unlabeled (\eg object boundaries in Pascal VOC, \emph{void} class in Cityscapes), and should be ignored during an attack. We use a binary mask $\bm{m}\in\{0,1\}^d$ to encode this, effectively reducing the number of constraints from $d$ to $\norm{\bm{m}}_1$.
Given the number of constraints considered in this problem (\eg ${\sim}10^6$ for Cityscapes), we consider an attack as successful if it satisfies at least a fraction $\nu$ of the constraints. In this paper, we use $\nu=99\%$.
To consolidate this in our attack, we compute a mask $\tilde{\vm}^{(t)}\in\{0,1\}^d$ at each iteration $t\in \{1,\ldots,N\}$ such that $\nu\norm{\vm}_1 \leq \|\tilde{\vm}^{(t)}\|_1 \leq \norm{\vm}_1$.
This mask discards the largest constraints, aligning the optimization objective with the criterion of successful attack. 
The discarded constraints are less likely to be satisfied, so this avoids the continuous increase of their associated multipliers, which may result in larger perturbations.
In practice, at each iteration $t$, we use a threshold $\xi^{(t)}$ as a percentile of the constraints $\vd^{(t)}$ to obtain $\tilde{\vm}^{(t)}$, and linearly decay $\xi^{(t)}$ from $100\%$ at the first iteration to $\nu$ at the last iteration:
\begin{equation}
\begin{aligned}
\label{eq:constraint_mask}
    \xi^{(t)} &= \Big(1 - (1-\nu)\frac{t - 1}{N - 1}\Big)\text{-percentile of }\vd^{(t)},\\
    \tilde{\vm}^{(t)} &= [\vd^{(t)} \leq \xi^{(t)}],
\end{aligned}
\end{equation}
where $[\cdot]$ is the Iverson bracket applied componentwise.

\paragraph{Constraint scaling} When dealing with large numbers of constraints in Augmented Lagrangian methods, it is standard practice to scale them (see section 12.5 of \cite{birgin2014practical}). However, there is no principled way of choosing this scale, as it is problem dependent. At iteration $t$, we multiply all the constraints by a scaling factor $w^{(t)}\in]0, 1]$ by computing
\begin{equation}
\label{eq:loss_alma}
    h_2(\pert) = (\tilde{\bm{m}}^{(t)})^\top P\big(w^{(t)}\bm{d}^{(t)}, \bm{\rho}^{(t)}, \bm{\mu}^{(t)}\big).
\end{equation}
This scaling factor is adjusted during the attack with a binary decision: if the pixel success rate at the current iteration is less than the target pixel success rate $\nu$, the scaling factor is increased, otherwise it is decreased:
\begin{equation}
    w^{(t)} = w^{(t-1)} \times \begin{cases}
        \frac{1}{1-\gamma_w} &\text{if} \quad \frac{\bm{m}^\top[\bm{d}^{(t)} \leq 0]}{\norm{\bm{m}}_1} < \nu;\\
        \frac{1}{1+\gamma_w} &\text{otherwise.}
    \end{cases}
\end{equation}
Multiplying by $\frac{1}{1\pm \gamma_w}$ instead of $1\pm\gamma_w$ biases $w^{(t)}$ to increase in case of oscillations, with two iterations resulting in a multiplication of $\frac{1}{1-\gamma_w^2}$ instead of $1-\gamma_w^2$.
$w^{(t)}$ is further projected on a safeguarding interval $[w_{\min}, 1]$. We use $w_{\min}=0.1$ and set the initial scale to $w^{(0)} = 1$.

\paragraph{Penalty} For the penalty function, we use a slightly modified version of the $P_2$ penalty from \cite{birgin2014practical}. It corresponds to the following mapping applied componentwise defined $(\forall y \in \mathbb{R})(\forall (\rho,\mu)\in [0,+\infty[^2)$:
\begin{equation}
\begin{gathered}
    P(y, \rho, \mu) = \begin{cases}
    \mu y + \mu \rho y^2 + \frac{1}{6}\rho^2 y^3 &\text{if} \quad y \geq 0;\\
    \frac{\mu y}{1 - \max(1, \rho)y} &\text{otherwise.}
    \end{cases}
\end{gathered}
\end{equation}
This penalty also satisfies the requirements in \cite{birgin2014practical}, while allowing the penalty multipliers to decrease faster for negative values of the constraint with $\rho < 1$.

\subsection{Proximal splitting}
\label{sec:proximal_splitting}

Proximal methods have been the workhorse of nonsmooth large scale optimization in the last two decades \cite{combettes2011proximal,bach2012optimization}. Originally designed for solving convex optimization problems, they have been also successfully employed in the nonconvex case \cite{attouch2013convergence}. One of their main advantages is that they offer the ability to split the cost function in a sum of terms which can be addressed either by computing their proximity operator, or their gradient when they are smooth. One important point is that, generally, the limitations on the step-size arising in proximal algorithms are related to the Lipschitz-regularity of the gradient of the smooth part, whereas the proximity operators of the other functions do not introduce any restriction. %

As a common practice in adversarial attacks, we are interested in minimizing the $\linf$-norm.
To handle this term, at each iteration of the proposed algorithm, our problem is expressed as the minimization of a sum of two functions as in \eqref{eq:alma_prox_loss}, where $h_1$ is convex and $h_2$ is smooth.
With this formulation, we can solve the problem using a \emph{forward-backward} splitting algorithm where the update reads
\begin{equation}
\label{eq:proximal_gradient_descent}
    \pert^{(t+1)} = \underbrace{\prox{\lambda h_1}}_{\text{backward step}}\big(\underbrace{\pert^{(t)} - \lambda \nabla_{\pert}h_2\big(\pert^{(t)}\big)}_{\text{forward step}}\big),
\end{equation}
$\lambda$ is a positive step-size, and $\prox{\lambda h_1}$ is the proximity operator of the function $\lambda h_1$. Global convergence guarantees of the forward-backward algorithm exist in the convex case \cite{combettes2005signal} and local ones in the nonconvex case \cite{attouch2013convergence}.
To carry out our attack, we thus need to find the proximity operator of $h_1$.

\paragraph{Proximity operator of $h_1$}

The function $h_1$ is a sum of two functions: $\norm{\cdot}_\infty$ and $\iota_\pertset$.
In general, the proximity operator of a sum of functions \emph{is neither} the sum of the proximity operators of the functions \emph{nor} their composition. One way to solve the sub-problem of finding the $\prox{}$ of $h_1 = \norm{\cdot}_\infty + \iota_\pertset$ would be to resort to an iterative proximal splitting algorithm, such as the dual forward-backward splitting \cite{combettes2011proximal}. However, we will propose a more efficient numerical approach by going back to the definition of the proximity operator \cite{bauschke2011convex}:
\begin{equation}
\label{eq:prox_infnite_indicator}
\begin{aligned}
    \prox{\lambda h_1}(\pert) &= \prox{\lambda \norm{\cdot}_\infty+\iota_\pertset}(\pert)\\
    &= \argmin_{\p\in \mathbb{R}^{Cd}} \frac12\norm{\p - \pert}_2^2 + \lambda \norm{\p}_\infty + \iota_\pertset(\p).
\end{aligned}
\end{equation}
As $\X=[0, 1]^{Cd}$, we can reformulate this problem as:
\begin{equation}
\label{eq:prox_problem_beta}
\begin{aligned}
    \min_{\p, \beta} \quad \frac12\norm{\p - \pert}_2^2 + \lambda \beta \quad
    \st \quad &{-}\beta \bm{1}_{Cd} \leq\p\leq\beta \bm{1}_{Cd}, \\
    &{-}\x\leq\p\leq \bm{1}_{Cd}-\x,
\end{aligned}
\end{equation}
where $\bm{1}_{Cd} = [1,\ldots,1]^\top \in \mathbb{R}^{Cd}$.
Since $\pertset=[0,1]^{Cd}-\x\subset [-1, 1]^{Cd}$,
the maximum possible $\linf$-norm of the solution $\p^\star$ is $1$, so we need to solve \eqref{eq:prox_problem_beta} for $\beta\in[0, 1]$. For $\beta = 0$, the solution is trivially
 $\p^\star=0$.
For $\beta=1$, the solution is $\p^\star = \proj{\pertset}(\pert)$ since the problem reduces to
\begin{equation}
    \min_{\p} \quad \frac12\norm{\p - \pert}_2^2 \quad \st \quad -\x\leq\p\leq \bm{1}_{Cd}-\x.
\end{equation}
For a given $\beta\in]0, 1[$,  \eqref{eq:prox_problem_beta} becomes a projection problem, so we can express the solution for $\p$ as $\p_\beta = \proj{[-\beta,\beta]^{Cd}\cap \pertset}(\pert)$.
The optimal value of $\beta$ has then to be found by solving
\begin{equation}
\label{eq:beta_problem}
    \min_{\beta\in [0,1]} \quad \frac12\norm{\p_\beta - \pert}_2^2 + \lambda \beta.
\end{equation}

\begin{proposition}
\label{prop:upper_bound_beta}
Problem \eqref{eq:beta_problem} is convex and admits a unique optimal solution $\beta^\star \in [0,1]$, for which the following inequality holds:
\begin{equation}
\label{e:boundbetastar}
    0 \leq \beta^\star \leq \norm{\proj{\pertset}(\pert)}_\infty.
\end{equation}
\end{proposition}
The proof of this result is provided in \autoref{sec:proof_upper_bound_beta}. Several methods can be used to solve \eqref{eq:beta_problem}. We employ a ternary search since it offers a linear convergence rate, avoids the computation of the gradient of the objective \wrt $\beta$, and is numerically stable. It also yields a number of iterations depending on the required absolute precision $\Delta \beta$. Based on inequality \eqref{e:boundbetastar}, the number of iterations for the ternary search is $\log(\nicefrac{\Delta \beta}{\norm{\proj{\pertset}(\pert)}_\infty}) / \log(\nicefrac23)$. Since we are performing adversarial attacks on images, the precision is usually 8-bit, or $\nicefrac{1}{255}\approx3.9\times10^{-3}$, we solve the problem with $\Delta \beta = 10^{-5}$, yielding 29 iterations at worse, when $\norm{\proj{\pertset}(\pert)}_\infty = 1$. The procedure to compute $\p^\star=\prox{\lambda\norm{\cdot}_{\infty}+\iota_\pertset}(\pert)$ is described in \autoref{alg:ternary_search}.

\begin{algorithm}
    \small
    \caption{Ternary search for $\p^\star=\prox{\lambda\norm{\cdot}_{\infty}+\iota_\pertset}(\pert)$}
    \label{alg:ternary_search}
    \begin{algorithmic}[1]
        \Require Input vector $\pert$ and feasible set $\pertset$.
        \Require Absolute precision $\Delta\beta$ on the solution.
        \State $\pert_\pertset \gets \proj{\pertset}(\pert)$
        \State $l \gets 0$ \Comment{Lower bound for $\beta^\star$}
        \State $u \gets \norm{\pert_\pertset}_\infty$ \Comment{Upper bound for $\beta^\star$}
        \State $n \gets \lceil\log(\nicefrac{\Delta\beta}{u}) / \log(\nicefrac23)\rceil$  \Comment{Number of steps}
        \For{$t \gets 1, \dots, n$}
            \State $\beta_l \gets l + (u - l) / 3$  \Comment{Points to evaluate the objective}
            \State $\beta_u \gets u - (u - l) / 3$
            \State $\p_l \gets \proj{[-\beta_l, \beta_l]}(\pert_\pertset)$  \Comment{Projection on interval}
            \State $\p_u \gets \proj{[-\beta_u, \beta_u]}(\pert_\pertset)$
            \State $f_l \gets \frac12 \norm{\p_l - \pert}_2^2 + \lambda \beta_l$  \Comment{Value of objective}
            \State $f_u \gets \frac12 \norm{\p_u - \pert}_2^2 + \lambda \beta_u$
            \If{$f_l \geq f_u$}  \Comment{Update bounds for search}
                \State $l \gets \beta_l$
            \Else
                \State $u \gets \beta_u$
            \EndIf
        \EndFor
        \State $\beta^\star = (l + u) / 2$  \Comment{Minimum is in $[l,u]$}
        \State \Return $\p^\star = \proj{[-\beta^\star, \beta^\star]}(\pert_\pertset)$  \Comment{Return $\prox{}$ using $\pert_\pertset$}
    \end{algorithmic}
\end{algorithm}

\paragraph{Variable Metric Forward-Backward} Forward-Backward algorithms can suffer from slow convergence. Therefore, we propose to use a Variable Metric Forward Backward (VMFB) \cite{chouzenoux2014variable, combettes2014variable} algorithm as an acceleration. 

\begin{definition*}
    Let $\metric$ be a positive definite matrix in $\R^{Cd\times Cd}$. For all $\lambda>0$, the proximity operator of $\lambda h_1$ in the metric $\metric$ is defined as
    \begin{equation}
        \prox{\lambda h_1}^\metric(\pert) = \argmin_{\p\in\R^{Cd}} \frac12 \norm{\p - \pert}_\metric^2 + \lambda h_1(\pert),
    \end{equation}
    where $\norm{\p}_\metric = \sqrt{\p^\top\metric \p}$ is a weighted norm.
\end{definition*}

Following the definition, the update in VMFB reads
\begin{equation}
\label{eq:vmfb}
    \pert^{(t+1)} = \prox{\lambda h_1}^\metric\left(\pert^{(t)} - \lambda \metric^{-1} \nabla_{\pert}h_2\big(\pert^{(t)}\big)\right).
\end{equation}
This method requires inverting a $Cd{\times}Cd$ square matrix, which may be impractical. Therefore, we use a diagonal metric $\metric=\operatorname{Diag}(\bm{s})$ with $\bm{s}\in ]0,+\infty[^{Cd}$. At iteration $t$, the diagonal vector $\bm{s}^{(t)}$ is estimated from the gradient as
\begin{equation}
\begin{aligned}
    \vv^{(t)} &= \alpha \vv^{(t-1)} + (1 - \alpha) \big(\nabla_{\pert}h_2(\pert^{(t)})\big)^2\\
    \bm{s}^{(t)} &= \sqrt{\frac{\vv^{(t)}}{1 - \alpha^t}} + \varepsilon
\end{aligned}
\end{equation}
where the square and square root operations are performed componentwise, $\alpha\in[0, 1[$ is a smoothing parameter, and $\vv^{(0)}=\bm{0}_{Cd}$.
\begin{remark}
    With this choice for the metric, the forward (\ie gradient) step becomes equivalent to the one in the Adam algorithm \cite{kingma2015adam} with $(\beta_1, \beta_2)=(0, \alpha)$.
\end{remark}

Note that \autoref{prop:upper_bound_beta} still holds for a diagonal metric. Indeed, the projection on $\pertset$ \wrt a diagonal metric $\proj{\pertset}^\metric = \argmin_{\vy\in\pertset} \norm{\cdot - \vy}_\metric^2$ is equal to the Euclidean projection $\proj{\pertset} = \argmin_{\vy\in\pertset} \norm{\cdot - \vy}_2^2$. This is readily deduced from the fact that both projection problems are separable in each component, since $\metric$ is diagonal and the constraint $\vy\in\pertset$ is separable. Therefore, since $\proj{\pertset}^\metric = \proj{\pertset}$, the proof is unchanged.
Additionally, the ternary search approach to compute $\p^\star$ can still be used: the norms in steps 10 and 11 are replaced by the weighted norm.

%% file: 5_experiments.tex
\section{Experiments}

\subsection{Ternary Search}

\begin{figure}
    \centering
    \includegraphics[width=\linewidth]{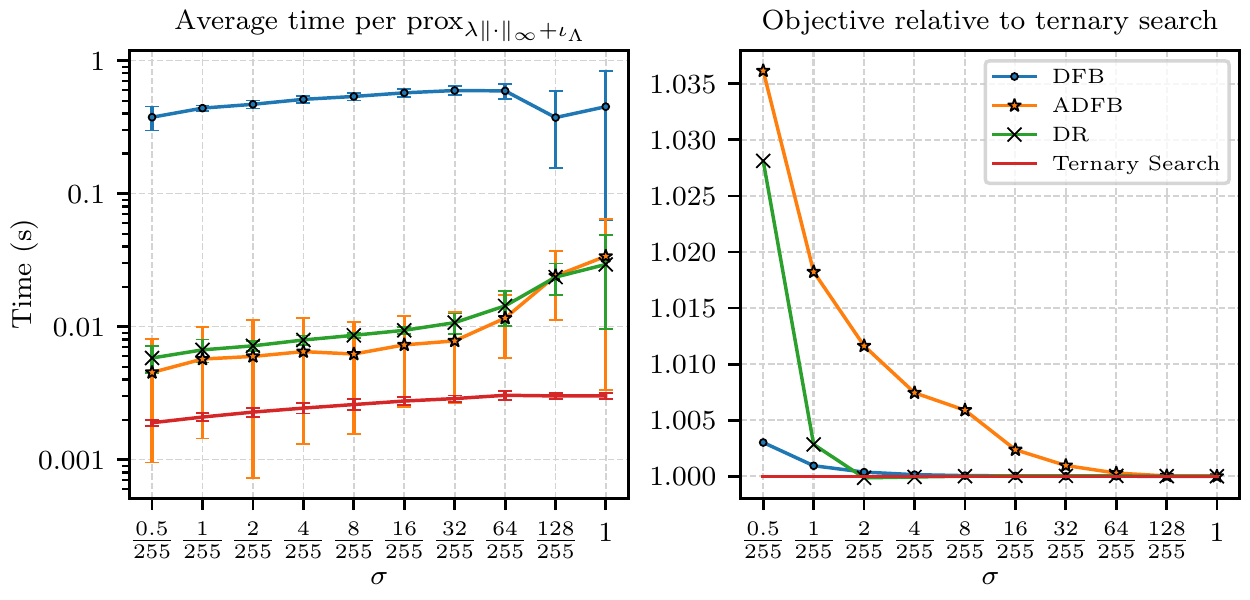}
    \caption{Comparison of average run-time and quality of solution in terms of optimization objective of DFB, ADFB, DR and Ternary Search for pseudorandom vectors $\pert\sim\mathcal{N}(0, \sigma^2 I_{d})$.}
    \label{fig:comparison_ternary_dfb_adfb}
\end{figure}

To evaluate the efficiency of our method to compute $\prox{\lambda h_1}$ using a ternary search, we provide a comparison with several traditional iterative splitting algorithms: Dual Forward-Backward (DFB) splitting \cite{combettes2010dualization}, an Accelerated variant (ADFB) using Nesterov acceleration on the dual problem \cite{chambolle2015convergence}, and a Douglas-Rachford (DR) splitting algorithm \cite{lions1979splitting}. We generate pseudorandom samples $\x\sim\mathcal{U}(\bm{0}_d,\bm{1}_d)$, perturbation vectors $\pert\sim\mathcal{N}(0, \sigma^2 I_{d})$, and scale $\lambda\sim 10^{\mathcal{U}(-1, 3)}$, with $d=2^{18}=512{\times}512$. For all methods, we use an absolute stopping criterion on the solution $\norm{\p^{(t+1)} - \p^{(t)}}_\infty \leq 10^{-5}$. For each $\sigma$, we repeat the evaluation 100 times, and report the average compute time\footnote{Experiments were run on NVIDIA A100 SXM4 $40$ GB GPUs.} and the value of the objective from \eqref{eq:prox_infnite_indicator} relative to the one obtained using the ternary search.

The results are shown in \autoref{fig:comparison_ternary_dfb_adfb}. Overall, the Ternary Search approach is faster and provides more accurate solutions to the computation of the proximity operator of interest than DFB, ADFB and DR. In some occasions, it may happen that ADFB converges faster, but the trade-off is a worse solution in terms of objective on average.

\subsection{Datasets and Models}

We perform experiments on the validation set of two well-known segmentation datasets: Pascal VOC 2012 \cite{pascal-voc-2012} and Cityscapes \cite{cordts2016cityscapes}. We evaluate the attacks on a collection of models chosen based on several criteria: widespread usage, high performance, and architectures diversity. With these criteria, we selected DeepLabV3+ ResNet-50, ResNet-101 \cite{chen2018encoder}, and FCN HRNetV2 W48 \cite{wang2020deep} which are CNN based models, and SegFormer MiT-B0 and MiT-B3 \cite{xie2021segformer} which are transformer based models.
We also consider the robust model DeepLabV3 ResNet-50 DDC-AT from \cite{xu2021dynamic}.

For white-box attacks, most attack algorithms compute the gradient of a loss \wrt the input. In the context of adversarial attacks, this quantity is computed for validation images, which can be larger than the image crops used in training. For Cityscapes specifically, models are evaluated on $2\,048{\times}1\,024$ images. This leads to high memory usage, and, in particular, DeepLabV3+ ResNet-101 and larger variants of the SegFormer family (\ie MiT-B4 and MiT-B5) require more than $40$ GB of memory per gradient computation on $2\,048{\times}1\,024$ images.
Therefore, we refrain from doing experiments on models requiring more than $40$ GB of GPU memory to ease future comparisons.

For most segmentation models, the evaluation protocol involves resizing the image to a specified size (\eg such that the smallest side has a specified length, while keeping the aspect ratio), using the model to produce a segmentation, and then resizing this segmentation to the original image size and compute the performance metrics. In our context, we do not perform this resizing before and after. This slight modification of the evaluation protocol leads to marginal differences in the typical performance metrics, which we report in \autoref{sec:model_performance}. All the models weights are fetched from the MMSegmentation library \cite{mmseg2020}, except for the robust model DeepLabV3 DDC-AT from \cite{xu2021dynamic}, which was obtained from the repository associated with the paper.

\subsection{Metrics}

In segmentation, the concept of a \textit{successful attack} is more ambiguous than in classification, where success is a binary criterion. From a security perspective in a segmentation task, \textit{mostly} making wrong predictions, except for a few pixels, can be considered as a successful attack (or a model failure), even though all the constraints are not satisfied.

Previous works on adversarial examples in a segmentation context \cite{arnab2018robustness,xie2017adversarial,xu2021dynamic} measure the model robustness (or equivalently, attack performance) using the mean Intersection over Union (mIoU) over all classes. However, this metric is biased towards small regions, and does not indicate how well the adversarial optimization problem \eqref{eq:minimal_adv_optim} is solved. Therefore, to measure the success of an attack, we simply measure the constraint satisfaction rate over all pixels in the mask $\bm{m}$, irrespective of the original class. For a given image, we call this constraint satisfaction rate the Attack Pixel Success Rate ($\mathrm{APSR}$). For untargeted attacks, the $\mathrm{APSR}$ is defined as:
\begin{equation}
    \mathrm{APSR} = \frac{\bm{m}^\top [\argmax_k f(\x+\pert)_{k,i} \neq \ytrue_i]}{\norm{\bm{m}}_1} \in [0,1]
\end{equation}
Although our approach has been presented in the context of untargeted attacks, it can also be straightforwardly extended to targeted attacks. In such a case, a target label $\ytarget = (\ytarget_i)_{1\le i \le d}$ is provided and the statement becomes $\argmax_k f(\x+\pert)_{k,i} = \ytarget_i$.
    
From there, choosing a specific threshold is arbitrary. In our experiments, we use $\nu=99\%$ as a threshold for the $\mathrm{APSR}$, meaning that an attack is considered successful if $\mathrm{APSR} \geq 99\%$. From an optimization perspective, this indicates that we satisfy at least $99\%$ of the constraints in problem \eqref{eq:minimal_adv_optim}. A lower threshold could also result in low quality segmentations. However, given the small $\linf$-norms of the perturbations observed to satisfy such a high threshold, we argue that it highlights the effectiveness of our attack.%

\subsection{Attack objectives}
\label{sec:attack_objective}

The most common scenario for adversarial attacks is the untargeted setting. While it is natural for classification, this can produce unrealistic segmentations that could easily be filtered or rejected (see \autoref{fig:untargeted_seg}). Therefore, we need to find a more natural objective for the adversarial attack depending on the context. For Pascal VOC, there is a \textit{background} class, which can be chosen as the target class for the whole image. This means that a successful targeted attack would produce a segmentation with no object of interest.

In contrast, Cityscapes does not consider a \textit{background} class, so we need a plausible target. A strategy explored in \cite{fischer2017adversarial} is to \textit{erase} a class and select the nearest neighbor that is not from the same class as the target label. This produces natural segmentation labels given the context, but it is not clear which class should be erased in general. For our experiments, we compute a target label based on the majority label for each pixel over the whole training set. This produces a more natural looking segmentation, but with high spatial frequencies, which differs from the dataset segmentation labels. Therefore, we draw a smoothed version of this target, avoiding the high frequencies, while keeping the structure. This target is provided in \autoref{sec:target_label}.

\subsection{Attacks}
\label{sec:attacks}

\begin{figure}
    \centering
    \includegraphics[width=\columnwidth]{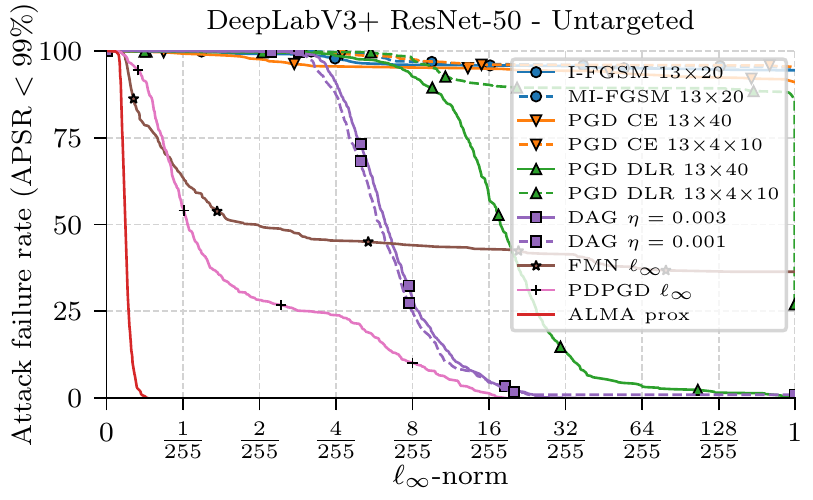}
    \caption{Percentage of unsuccessful untargeted attacks for DeepLabV3+ ResNet-50 on Cityscapes. Horizontal axis is linear on [0, \nicefrac{2}{255}] and logarithmic on [\nicefrac{2}{255}, 1].}
    \label{fig:deeplabv3plus_cityscapes_untargeted}
    \vspace{-3mm}
\end{figure}

For the minimization attacks, we set a 500 iterations budget, corresponding to a ${\sim}24$ hours run-time to attack the entire validation set of Cityscapes with the largest model.%

\paragraph{DAG}
The main baseline in our experiments is DAG \cite{xie2017adversarial}.
We report results with two step-sizes (0.003 and 0.001) such that the larger does not fail to find adversarial perturbations, while the smaller finds smaller perturbations for some samples, but fails on others.

\paragraph{Other baselines} 
As mentioned in \autoref{sec:related_works}, we can adapt some attacks originally designed for classification.
We consider I-FGSM \cite{kurakin2017adversarial} and MI-FGSM \cite{dong2018boosting} with 20 iterations and use a 13 steps binary search on the $\linf$-norm, yielding an error of $2^{-13}\approx 10^{-4}$, to find the smallest norm for which the attack succeeds. Similarly, we include four variants of the PGD attack \cite{madry2018towards} with a binary search. These variants use the Cross-Entropy (CE) or the DLR loss from \cite{croce2020reliable}, in combination with 40 steps or 10 steps with 3 random restarts (totaling 40 steps).\footnote{Given the poor performance of these attacks on the regularly trained models, we do not evaluate them on the robust model from \cite{xu2021dynamic}.}
We also adapt
the $\linf$ variant of FMN \cite{pintor2021fast} by taking the loss as the average over the mask $\vm$ and testing if $\mathrm{APSR}\geq \nu$ as the binary decision for the projection step.
We use $\alpha=10$ in FMN (see section 3.1 of \cite{pintor2021fast}). 
Finally, we modify the PDPGD attack \cite{matyasko2021pdpgd} to take into account the constraint masking strategies. We provide the details of the modifications, as well as an ablation study on the impact of these strategies in \autoref{sec:details_pdpgd_alma}. We could not include the Houdini attack \cite{cisse2017houdini} as no public implementation is available.

\paragraph{ALMA prox} For our attack, we use an initial step-size $\lambda^{(0)}=10^{-3}$ and decay it to $\lambda^{(N)}=10^{-4}$. With a budget of 500 iterations, we set $\alpha=0.8$, compared to 0.9 for 1\,000 iterations in \cite{rony2021augmented}. We set the $\mu^{(0)}=\bm{1}_d$ and $\rho^{(0)}=0.01\cdot\bm{1}_d$, the penalty parameter increase rate $\gamma=2$ and the constraint scale adjustment rate to $\gamma_w=0.02$.

%% file: 6_results.tex
\section{Results}
\label{sec:results}

\begin{table*}
    \footnotesize
    \centering
    \begin{tabular}{llrrcrrcrrcrr}
        & & \multicolumn{5}{c}{Pascal VOC 2012} & & \multicolumn{5}{c}{Cityscapes} \\
        \cmidrule{3-7}\cmidrule{9-13}
        & Attack & \multicolumn{2}{c}{\makecell{DeepLabV3+\\ResNet-50}} & &  \multicolumn{2}{c}{\makecell{FCN\\ HRNetV2 W48}} & & \multicolumn{2}{c}{\makecell{DeepLabV3+\\ResNet-50}} & & \multicolumn{2}{c}{\makecell{SegFormer\\MiT-B3}} \\
        \midrule
        \multirow{11}{*}{\begin{turn}{90}Untargeted\end{turn}}
        & I-FGSM $13{\times}20$ \cite{kurakin2017adversarial}        & 146.32 & 136.55 & & 227.11 & 145.69 &  & 255.00 & 242.15 & & 255.00 & 208.97 \\
        & MI-FGSM $13{\times}20$ \cite{dong2018boosting}             & 195.35 & 145.96 & & 255.00 & 157.63 &  & 255.00 & 244.43 & & 255.00 & 228.43 \\
        & PGD CE $13{\times}40$ \cite{madry2018towards}              &  80.10 & 120.93 & & 152.08 & 136.05 &  & 255.00 & 236.77 & & 255.00 & 188.15 \\
        & PGD CE $13{\times}4{\times}10$ \cite{madry2018towards}      &  24.90 &  74.15 & &  66.93 & 118.76 &  & 255.00 & 244.92 & & 255.00 & 231.51 \\
        & PGD DLR $13{\times}40$ \cite{madry2018towards}             &   7.22 &  26.42 & &  11.11 &  23.85 &  &  17.75 &  24.23 & &  84.22 & 102.80 \\
        & PGD DLR $13{\times}4{\times}10$ \cite{madry2018towards}     &   4.42 &  24.99 & &  10.46 &  29.75 &  & 255.00 & 227.89 & & 110.82 & 134.41 \\
        & DAG $\eta=0.003$ \cite{xie2017adversarial}                &   5.69 &   6.63 & &   8.80 &  10.61 &  &   6.30 &   7.51 & &   8.83 & 9.02 \\
        & DAG $\eta=0.001$ \cite{xie2017adversarial}                &   5.23 &   8.22 & &   8.49 &  14.61 &  &   5.95 &   9.39 & &   8.59 & 53.65 \\
        & FMN $\linf$ \cite{pintor2021fast}                         &   0.46 &  38.34 & &   0.91 &  46.39 &  &   1.97 &  96.57 & &   1.08 & 6.42 \\
        & PDPGD $\linf$ \cite{matyasko2021pdpgd}                    &   0.73 &   1.77 & &   1.52 &   2.43 &  &   1.06 &   2.82 & &   1.39 & 3.05 \\
        & ALMA $\mathrm{prox}$                                      & \textbf{0.32} & \textbf{0.34} & & \textbf{0.51} & \textbf{0.56} & & \textbf{0.24} & \textbf{0.26} & & \textbf{0.33} & \textbf{0.33} \\
        \midrule
        \multirow{11}{*}{\begin{turn}{90}Targeted\end{turn}}
        & I-FGSM $13{\times}20$ \cite{kurakin2017adversarial}        & 0.47 & 0.50 & & 0.59 & 0.64 &  & 255.00 & 255.00 & &  51.73 & 88.24 \\
        & MI-FGSM $13{\times}20$ \cite{dong2018boosting}             & 0.59 & 0.66 & & 0.77 & 0.85 &  &   4.26 &  55.35 & &   2.54 & 2.60 \\
        & PGD CE $13{\times}40$ \cite{madry2018towards}              & 0.37 & 0.43 & & 0.50 & 0.54 &  &   3.49 &   3.71 & &   1.87 & 1.92 \\
        & PGD CE $13{\times}4{\times}10$ \cite{madry2018towards}      & 0.50 & 0.51 & & 0.62 & 0.65 &  & 255.00 & 255.00 & & 255.00 & 255.00 \\
        & PGD DLR $13{\times}40$ \cite{madry2018towards}             & 0.62 & 0.68 & & 0.81 & 0.94 &  &   8.37 &  67.86 & &   3.98 & 4.09 \\
        & PGD DLR $13{\times}4{\times}10$ \cite{madry2018towards}     & 0.87 & 1.07 & & 1.12 & 1.28 &  & 255.00 & 255.00 & & 255.00 & 255.00 \\
        & DAG $\eta=0.003$ \cite{xie2017adversarial}                & 4.21 & 4.50 & & 5.32 & 5.66 &  &  11.34 &  12.96 & &   9.82 & 10.06 \\
        & DAG $\eta=0.001$ \cite{xie2017adversarial}                & 3.92 & 4.21 & & 5.07 & 5.36 &  &  10.96 &  40.28 & &  11.53 & 119.47 \\
        & FMN $\linf$ \cite{pintor2021fast}                         & 0.42 & 0.45 & & 0.47 & 0.49 &  & 255.00 & 254.36 & & 255.00 & 255.00 \\
        & PDPGD $\linf$ \cite{matyasko2021pdpgd}                    & 0.28 & 0.36 & & 0.35 & 0.46 &  &  14.51 &  14.41 & &  19.26 & 19.20 \\
        & ALMA $\mathrm{prox}$                                      & \textbf{0.25} & \textbf{0.26} & & \textbf{0.32} & \textbf{0.34} & & \textbf{1.15} & \textbf{1.17} & & \textbf{0.65} & \textbf{0.66} \\
        \bottomrule
    \end{tabular}
    \caption{Median and average norms $\norm{\pert}_\infty{\times}255$ for each adversarial attack on Pascal VOC 2012 and Cityscapes.}
    \label{tab:partial_median_average_pert}
    \vspace{-3mm}
\end{table*}

\begin{table}
    \scriptsize
    \centering
    \begin{tabular}{llrrrr}
        & Attack & \multicolumn{2}{c}{Pascal VOC 2012} & \multicolumn{2}{c}{Cityscapes} \\
        \midrule
        \multirow{5}{*}{\begin{turn}{90}Untargeted\end{turn}}
        & DAG $\eta=0.003$ \cite{xie2017adversarial}                & 12.05             & 25.96             & 16.99      & 25.81      \\
        & DAG $\eta=0.001$ \cite{xie2017adversarial}                & 11.71             & 57.93             & 17.02      & 81.30      \\
        & FMN $\linf$ \cite{pintor2021fast}                         & 1.28              & 75.66             & 37.54      & 128.67     \\
        & PDPGD $\linf$ \cite{matyasko2021pdpgd}                    & 3.53              & 11.86             & 43.61      & 58.10      \\
        & ALMA $\mathrm{prox}$                                      & \textbf{0.78}     & \textbf{2.17}     & \textbf{0.81} & \textbf{1.01} \\
        \midrule
        \multirow{5}{*}{\begin{turn}{90}Targeted\end{turn}}
        & DAG $\eta=0.003$ \cite{xie2017adversarial}                & 7.01              & 7.68              & 31.60      & 69.84      \\
        & DAG $\eta=0.001$ \cite{xie2017adversarial}                & 6.78              & 8.78              & 255.00     & 230.29     \\
        & FMN $\linf$ \cite{pintor2021fast}                         & 1.31              & 36.10             & 255.00     & 255.00     \\
        & PDPGD $\linf$ \cite{matyasko2021pdpgd}                    & 0.84              & 1.23              & 255.00     & 255.00     \\
        & ALMA $\mathrm{prox}$                                      & \textbf{0.52}     & \textbf{0.54}     & \textbf{3.64} & \textbf{3.92} \\
        \bottomrule
    \end{tabular}
    \caption{Median and average norms $\norm{\pert}_\infty{\times}255$ for each adversarial attack on the robust model DeepLabV3 DDC-AT \cite{xu2021dynamic}.}
    \label{tab:median_average_pert_ddcat}
    \vspace{-3mm}
\end{table}

\paragraph{Perturbation size}

For each dataset, model, and scenario (untargeted and targeted), we plot the attack failure rate as a function of the perturbation size. This failure rate corresponds to the fraction of samples for which an attack has \textit{not} found an adversarial example with $\mathrm{APSR}\geq99\%$. This kind of plot can be interpreted in two ways: model-centric and attack-centric. In a model-centric analysis, a more robust model will require larger perturbations to be fooled, and therefore, correspond to a curve towards the upper right. In an attack-centric analysis, a stronger attack will find smaller perturbations, and have a curve towards the lower left.

\autoref{fig:deeplabv3plus_cityscapes_untargeted} shows this plot for untargeted attacks on Cityscapes with DeepLabV3+ ResNet-50. To improve readability, we use a linear scale on $[0, \nicefrac{2}{255}]$ and a logarithmic scale on  $[\nicefrac{2}{255}, 1]$. Here, DAG needs a norm of $\nicefrac{8}{255}$ to successfully fool ${\sim}75\%$ of the samples (with $\mathrm{APSR}\geq99\%$), and $\nicefrac{24}{255}$ to fool 99\% of the samples. In contrast, ALMA $\prox{}$ finds adversarial perturbations with $\linf$-norms smaller than $\nicefrac{0.55}{255}$ for \emph{all} samples. This contradicts the robustness results in \cite{arnab2018robustness}, where models have mIoUs of ${\sim}50\%$ at $\nicefrac{1}{255}$, as opposed to $2.1\%$ at $\nicefrac{0.55}{255}$ here.

\autoref{tab:partial_median_average_pert} reports the median and average $\linf$-norm (multiplied by 255 for readability) of the perturbations produced by the attacks for a subset of the regular models: DeepLabV3+ ResNet-50 and FCN HRNetV2 W46 on Pascal VOC 2012, and DeepLabV3+ ResNet-50 and SegFormer MiT-B3 on Cityscapes. For unsuccessful attacks (\ie $\mathrm{APSR} < 99\%$), the perturbation size is considered to be 1 (\ie $\nicefrac{255}{255}$).
Overall, the ALMA $\prox{}$ attack outperforms all the attacks considered in our experiments. On Pascal VOC, most attacks can find small (\eg $\leq \nicefrac{1}{255}$) perturbations in the targeted scenario. This is expected because the most frequent class in Pascal VOC is the background. However, the differences are much larger in the untargeted scenario, which corresponds to predicting mostly non-background classes. On Cityscapes, the scale of the problem (${\sim}10^6$ constraints) highlights the difficulty of generating adversarial perturbations for segmentation models. PDPGD handles the untargeted scenario well, with median $\linf$-norms of ${\sim}\nicefrac{1}{255}$, but produces much larger perturbations in the targeted case. Contrarily, PDG CE $13{\times}40$ finds small targeted perturbations, but fails in the untargeted scenario. Finally, FMN has inconsistent performance across samples: on Cityscapes in the untargeted scenario, the low median and high average indicates that it fails to find for a large portion of the samples. This can also be seen in \autoref{fig:deeplabv3plus_cityscapes_untargeted}, where FMN finds perturbation with $\linf$-norms smaller than $\nicefrac{2}{255}$ for ${\sim}50\%$ of the samples, but fails for ${\sim}35\%$ of them. The results for all models can be found in \autoref{sec:attack_results} with similar trends on the other models.

Finally, the median and average perturbation norms for the robust model DeepLabV3 DDC-AT are reported in \autoref{tab:median_average_pert_ddcat}. On Pascal VOC, several attacks succeed in finding small perturbations. However, Cityscapes is, again, more challenging, especially in the targeted scenario. 

\paragraph{Attack complexity}

The observed average complexities in terms of number of forward and backward propagations of the model (see Section 4 of \cite{rony2021augmented}) and run-times are reported in \autoref{sec:attack_complexities}. 
The number of propagations is equal to the iteration budget for all attacks except DAG, which uses an early stopping criterion.
Therefore, DAG has lower complexity on average. For smaller step-sizes, it can reach the limit of 500 iterations for some samples (\eg for SegFormer models on Cityscapes, see \autoref{fig:linf_cityscapes}) and fails to find adversarial perturbations. 
The second observation is that the proximity operator used in ALMA $\prox{}$ does not significantly increase the run-time compared to the other attacks with similar number of forward and backward propagations. 

%% file: 7_conclusion.tex
\section{Conclusion}
\label{sec:conclusion}

We proposed an adversarial attack for deep semantic segmentation models to produce minimal perturbations \wrt the $\linf$-norm. Our attack is based on an Augmented Lagrangian method, which allows us to tackle large numbers of misclassification constraints using gradient-based optimization, coupled with a proximal splitting of the objective to minimize the $\linf$-norm and satisfy the input space constraints with a proximity operator. Additionally, we devised an efficient method to compute this proximity operator, which is compatible with a VMFB acceleration based on a diagonal metric. Our attack offers significant improvements in terms of $\linf$-norm minimization for segmentation tasks, even for a robust model. One limitation of our method is that it does not produce valid digitized images, \ie encoded with a reduced number of bits, such as 8. For a discussion on this issue, see \cite{bonnet2020what}. Note that all the attacks considered in our experiments do not produce valid images as well.

%% file: 9_appendix.tex
\appendix
\section{Dense Adversary Generation attack}
\label{sec:dag_attack}

In this section, we provide the algorithm of the Dense Adversary Generation (DAG) attack from \cite{xie2017adversarial}.
The main difference with the original algorithm proposed in \cite{xie2017adversarial} is the stopping criterion based on the pixel success rate in steps 9 to 11. In the original method published in \cite{xie2017adversarial}, the attack is supposed to stop once all the constraints are satisfied (\ie all pixels in the mask $\vm$ are adversarial). However, since this criterion is rarely satisfied, even after hundreds of iterations on a dataset like Cityscapes, the actual implementation made available in \url{https://github.com/cihangxie/DAG} uses the stopping criterion described in \autoref{alg:dag_attack}: the attack stops once a threshold of pixel success rate set to $99\%$ is reached. The threshold value used is thus identical to the one used in the experiments of this paper.

\begin{algorithm}
    \small
    \caption{DAG attack}
    \label{alg:dag_attack}
    \begin{algorithmic}[1]
        \Require Classifier $f$, original image $\x\in[0, 1]^{C\times H\times W}$, true or target label $\ytrue\in\mathbb{N}^{H\times W}$, binary mask $\vm\in\{0, 1\}^{H\times W}$
        \Require Step size $\eta$, maximum number of iterations $N$, threshold of pixel success rate $\nu$
        \State Initialize $\pert^{(0)} \gets \bm{0}$
        \State If targeted attack: $\mu\gets -1$ else $\mu\gets 1$
        \For{$t \gets 1, \dots, N$}
            \State $\xadv^{(t)} \gets \proj{[0, 1]}(\x + \pert^{(t-1)})$  \Comment{$\in[0, 1]^{C\times H\times W}$}
            \State $\z \gets f(\xadv^{(t)})$ \Comment{$\in\R^{K\times H\times W}$}
            \For{$i \gets 1, \dots, d$}
                \State $\bm{\Delta z}_i = \mu (\z_{\ytrue_i,i} - \max\limits_{j\neq \ytrue_i}\z_{j,i})$  \Comment{Difference of logits}
            \EndFor
            \State $r \gets \frac{\vm^\top[\bm{\Delta z} < 0]}{\norm{\vm}_1}$ \Comment{Pixel success rate}
            \If{$r \geq \nu$}
                \State \Return $\xadv^{(t)}$  \Comment{Stop the attack}
            \EndIf
            \State $\sL \gets \vm^\top \max\{0, \bm{\Delta z}_i\}$
            \State $\g \gets \nabla_{\pert}\sL$
            \State $\pert^{(t)} \gets \pert^{(t-1)} - \frac{\eta}{\norm{\g}_\infty}\g$  \Comment{Normalized gradient step}
        \EndFor
    \end{algorithmic}
\end{algorithm}

\section{Proof of Proposition 1}
\label{sec:proof_upper_bound_beta}

\begin{proof}
Problem \eqref{eq:prox_problem_beta} amounts to minimizing
function
\begin{multline}
    \Phi\colon (\p,\beta)\mapsto \frac12\norm{\p - \pert}_2^2 + \lambda \beta
    +\iota_{[0,+\infty[^{Cd}}(\beta \bm{1}_{Cd}-\p)\\
    +\iota_{[0,+\infty[^{Cd}}(\p+\beta \bm{1}_{Cd})
    + \iota_{\Lambda}(\p),
\end{multline}
defined on $\mathbb{R}^{Cd}\times \mathbb{R}$.
This function is convex since it is a sum of elementary convex functions of $(\p,\beta)$. It follows that its marginal function
\begin{equation}
    \underline{\Phi}\colon \beta \mapsto \inf_{\p\in \mathbb{R}^{Cd}} \Phi(\p,\beta)
\end{equation}
is convex. Since, for any given $\beta\in [0,1]$, 
$\p \mapsto \Phi(\p,\beta)$ is strongly convex and proper, it admits a unique minimizer $\p_\beta$. More precisely, we have, for every $\beta \in \mathbb{R}$,
\begin{equation}
\underline{\Phi}(\beta) = 
\begin{cases}
\frac12\norm{\p_\beta - \pert}_2^2 
+ \lambda \beta & \mbox{if $\beta \in [0,1]$}\\
+\infty & \mbox{otherwise.}
\end{cases}
\end{equation}
The above function admits a unique minimizer $\beta^\star \in [0,1]$
since $\p^\star=\p_{\beta^\star}$ is uniquely defined (as it is the proximity point of a proper lower-semicontinuous convex function) and $\beta^\star = \|\p_{\beta^\star}\|_{\infty}$.

Let $\p^\star = \prox{\lambda\norm{\cdot}_\infty + \iota_\pertset}(\pert)$ be the minimizer of \eqref{eq:prox_problem_beta} and let $\pert_\pertset = \proj{\pertset}(\pert)$. We have
\begin{equation}
     \frac12 \norm{\p^\star - \pert}_2^2 + \lambda \norm{\p^\star}_\infty \leq \frac12 \norm{\pert_\pertset - \pert}_2^2 + \lambda \norm{\pert_\pertset}_\infty
\end{equation}
Since $\pert_\pertset = \proj{\pertset}(\pert) = \argmin\limits_{\y\in\pertset} \norm{\y - \pert}_2^2$, we also have
\begin{equation}
\label{eq:proj_inequality}
    \norm{\pert_\pertset - \pert}_2^2 \leq \norm{\p^\star - \pert}_2^2
\end{equation}
Thus
\begin{equation}
\begin{aligned}
    \lambda \norm{\p^\star}_\infty &\leq \underbrace{\frac12(\norm{\pert_\pertset - \pert}_2^2 - \norm{\p^\star - \pert}_2^2)}_{\leq 0} + \lambda \norm{\pert_\pertset}_\infty \\
    &\leq \lambda \norm{\pert_\pertset}_\infty
\end{aligned}
\end{equation}
We deduce that $\boxed{\beta^\star = \norm{\p^\star}_\infty \leq \norm{\pert_\pertset}_\infty}$.
\end{proof}

\newpage
\section{ALMA prox attack algorithm}
\label{sec:alma_prox}

\begin{algorithm}
    \small
    \caption{ALMA $\prox{}$ attack (untargeted)}
    \label{alg:seg_alma_attack}
    \begin{algorithmic}[1]
        \Require Classifier $f$, original image $\x\in[0, 1]^{C\times H\times W}$, true or target label $\ytrue\in\mathbb{N}^{H\times W}$, binary mask $\vm\in\{0, 1\}^{H\times W}$
        \Require Threshold of pixel success rate $\nu$
        \Require Penalty function $P$, initial multiplier $\bm{\mu}^{(0)}\in\R_{++}^{H\times W}$, initial penalty parameter $\bm{\rho}^{(0)}\in\R_{++}^{H\times W}$ 
        \Require Minimum scale $w_{\min}$, scale adjustment rate $\gamma_w>0$
        \Require Number of iterations $N$, initial step size $\lambda^{(0)}$, penalty parameter increase rate $\gamma > 1$, constraint improvement rate $\tau$, $M$ number of steps between $\rho$ increase, $\alpha$ smoothing parameter
        \State Initialize $\pert^{(0)} \gets \bm{0}$, $\vv^{(0)} \gets 0$, $w^{(0)} \gets 1$
        \For{$t \gets 1, \dots, N$}
            \State $\xadv^{(t)} \gets \x + \pert^{(t-1)}$  \Comment{$\in[0, 1]^{C\times H\times W}$}
            \State $\bm{d}^{(t)} \gets \mathrm{DLR^+}(f(\xadv^{(t)}),\ytrue)$  \Comment{$\in\R^{H\times W}$}
            \If{$\frac{\vm^\top[\bm{d}^{(t)} \leq 0]}{\norm{\vm}_1} < \tau$} \Comment{Adjust constraint scale}
                \State $\hat{w} \gets \frac{w^{(t-1)}}{1-\gamma_w}$ \Comment{Increase scale}
            \Else
                \State $\hat{w} \gets \frac{w^{(t-1)}}{1+\gamma_w}$ \Comment{Decrease scale}
            \EndIf
            \State $w^{(t)} \gets \proj{[w_{\min},1]}(\hat{w})$
            
            \State $\xi^{(t)} \gets (1 - (1 - \nu)\frac{t - 1}{N - 1})\text{-percentile of }\bm{d}^{(t)}$
            \State $\tilde{\vm}^{(t)} \gets [\bm{d}^{(t)} \leq \xi^{(t)}]$ \Comment{$\in\{0,1\}^{H\times W}$}
            \State $\hat{\bm{\mu}} \gets \nabla_{\bm{d}} \left((\tilde{\vm}^{(t)})^\top P(w^{(t)}\bm{d}^{(t)},\bm{\rho}^{(t-1)},\bm{\mu}^{(t-1)})\right)$
            \State $\bm{\mu}^{(t)} \gets \proj{[\mu_\text{min},\mu_\text{max}]}\big(\alpha\bm{\mu}^{(t-1)}+(1-\alpha)\hat{\bm{\mu}}\big)$  \Comment{$\in\R_{++}^{H\times W}$}
            \For{$i\gets 1, \dots, d$} \Comment{$\rho$ adjustment}
                \If{$t\bmod M =0$\; \textbf{and} \; $\tilde{\vm}^{(t)}_i$ = 1 \; \textbf{and} \big($\exists j\in\{0, \dots, M-1\} : \bm{d}_i^{(t-j)} \leq 0$\; \textbf{or} \;$\bm{d}_i^{(t)}\leq\tau \bm{d}_i^{(t-M)}$\big)}
                    \State $\bm{\rho}_i^{(t)} \gets \bm{\rho}_i^{(t-1)}$ \Comment{Constraint improved or satisfied}
                \Else   
                    \State $\bm{\rho}_i^{(t)} \gets \gamma \bm{\rho}_i^{(t-1)}$
                \EndIf
            \EndFor
            \State $\sL \gets (\tilde{\vm}^{(t)})^\top P(w^{(t)} \bm{d}^{(t)}, \bm{\rho}^{(t)}, \bm{\mu}^{(t)})$  \Comment{$\in\R$}
            \State $\g^{(t)} \gets \nabla_{\pert}\sL$  \Comment{$\in\R^{C\times H\times W}$}
            \State $\vv^{(t)} \gets \alpha \vv^{(t-1)} + (1-\alpha) (\g^{(t)})^2 $  \Comment{$\in\R_{+}^{C\times H\times W}$}
            \State $\metric \gets \mathrm{Diag}\left(\sqrt{\frac{\vv^{(t)}}{1-\alpha^t}} + \varepsilon\right)$
            \State $\pert^{(t)} \gets \prox{\lambda^{(t)}\norm{\cdot}_\infty+\iota_\pertset}^{\metric}(\pert^{(t-1)} - \lambda^{(t)} \metric^{-1} \bm{g})$ \Comment{VMFB}
        \EndFor
        \State \Return $\xadv^{(t)}$ that is adversarial and has the smallest norm
    \end{algorithmic}
\end{algorithm}

\section{Image size and performance of the models}
\label{sec:model_performance}

For all models, except DeepLabV3 DDC-AT, the images of Pascal VOC 2012 are resized so that the smaller side is of length $512$ while keeping the aspect ratio, and for Cityscapes, the images keep their original size of $2\,048{\times}1\,024$.
For DeepLabV3 DDC-AT from \cite{xu2021dynamic}, the images of Pascal VOC 2012 are resized so that the longer side is of length 512 while keeping the aspect ratio, and for Cityscapes, the images are resized to $1\,024\times 512$.

\begin{table}
    \scriptsize
    \centering
    \resizebox{\columnwidth}{!}{
    \begin{tabular}{clcc}
    Dataset & Model & mIoU (\%) & \makecell{Pixel\\Accuracy (\%)} \\
    \midrule[0.75pt]
    \multirow{4}{*}{\makecell{Pascal VOC\\2012 (+Aug)}}
        & DeepLabV3+ ResNet-50 \cite{chen2018encoder} & $77.4_{+0.8}$ & $94.9_{+0.2}$\\
        & DeepLabV3+ ResNet-101\cite{chen2018encoder} & $78.8_{+0.1}$ & $95.3_{+0.1}$\\
        & FCN HRNetV2 W48 \cite{wang2020deep} & $76.4_{+0.2}$ & $94.7_{+0.1}$\\
        \cmidrule{2-4}
        & DeepLabV3 DDC-AT \cite{xu2021dynamic} & $75.2_{+0.0}$ & $94.4$ \\
    \midrule[0.75pt]
    \multirow{5}{*}{Cityscapes}
        & DeepLabV3+ ResNet-50 \cite{chen2018encoder} & $80.1_{-0.2}$ & $96.4_{-0.1}$\\
        & FCN HRNetV2 W48 \cite{wang2020deep} & $80.5_{-0.2}$ & $96.6_{-0.1}$\\
        & SegFormer MiT-B0 \cite{xie2021segformer} & $76.4_{-0.1}$ & $95.9_{-0.0}$\\
        & SegFormer MiT-B3 \cite{xie2021segformer} & $81.8_{-0.0}$ & $96.7_{-0.1}$\\
        \cmidrule{2-4}
        & DeepLabV3 DDC-AT \cite{xu2021dynamic} & $71.0_{-0.3}$ & $95.0$ \\
    \bottomrule
    \end{tabular}
    }
    \caption{Performance of the models used in the experiments on the validation sets. Numbers were obtained from our evaluation; subscripts correspond to the difference with the original evaluation protocol. For DeepLabV3 DDC-AT from \cite{xu2021dynamic}, the pixel accuracy was not reported.}
    \label{tab:models}
\end{table}

\section{Cityscapes target label}
\label{sec:target_label}

\begin{figure}[h]
    \centering
    \includegraphics[width=\columnwidth]{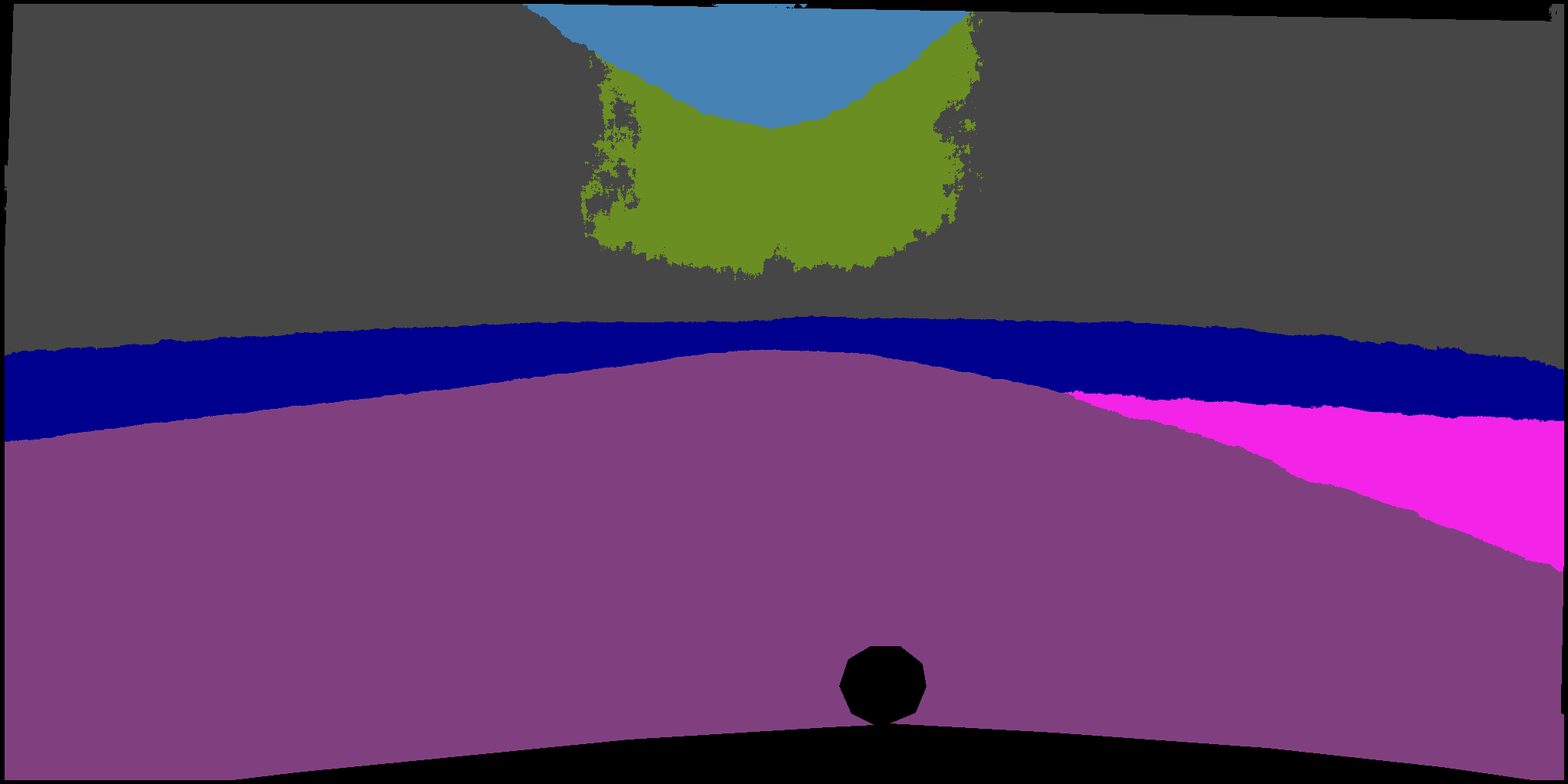}\\
    \vspace{0.5em}
    \includegraphics[width=\columnwidth]{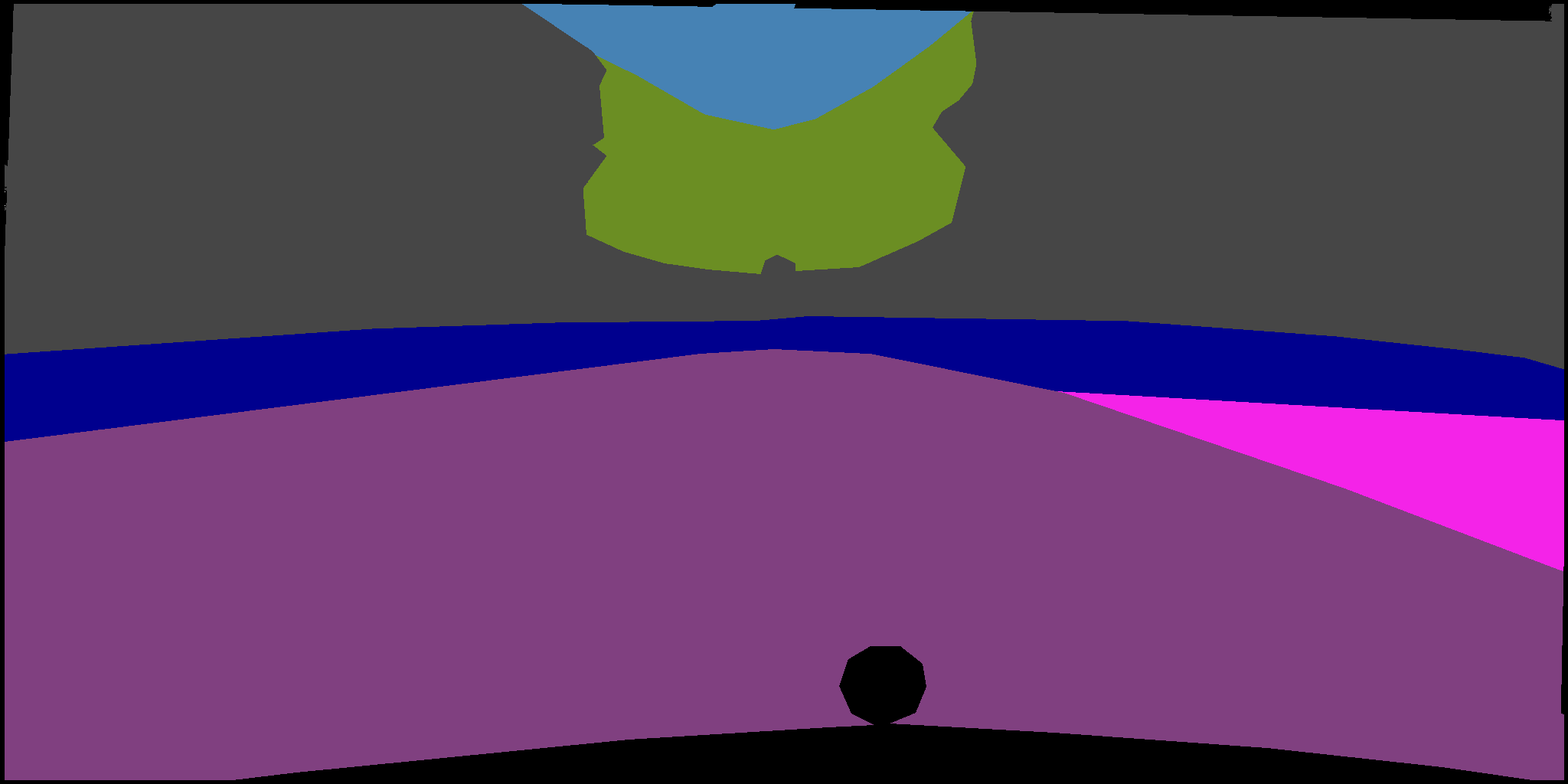}
    \caption{Cityscapes target segmentation used in our experiments. Top is the pixel majority label, bottom is the smoothed version. Classes in the target are: \textit{road} (purple), \textit{sidewalk} (pink), \textit{car} (blue), \textit{building} (dark gray), \textit{vegetation} (green), \textit{sky} (cyan) and \textit{no label} (black).}
    \label{fig:cityscapes_target}
\end{figure}

\newpage
\section{Masking strategies for ALMA prox and PDPGD}
\label{sec:details_pdpgd_alma}

In this section, we detail the modifications brought to PDPGD \cite{matyasko2021pdpgd} and study the effect of the masking strategies on PDPGD and ALMA prox. Besides the addition of perturbation tracking logic (\ie monitoring the best smallest perturbations during the optimization), we also incorporate the masking needed for the unlabeled regions.

\paragraph{Modifications for PDPGD}

For PDPGD, the dual variable $\lambda\in\R$ is replaced by a vector $\vlambda\in \R^d$. In \cite{matyasko2021pdpgd}, the gradient ascent on the dual variable is performed in the log domain, and is projected onto the 1-simplex (section IV of \cite{matyasko2021pdpgd}). For the segmentation variant, this translates into adding 1 to the denominator of the $\mathrm{softmax}$ function used to project onto the $(d-1)$-simplex. This is equivalent to padding $\vlambda$ with a 0 and projecting it on the $d$-simplex $\Delta^d\subset [0,1]^{d+1}$. The weight of the norm $\norm{\pert}$ in Equation (10) of \cite{matyasko2021pdpgd} becomes $1$ minus the sum of the weights of the constraints. Adding the mask $\vm$, this results in the following computations (introducing variables not described in \cite{matyasko2021pdpgd}):
\begin{equation}
\label{eq:pdpgd_loss}
\begin{aligned}
    \vlambda_\Delta &= \frac{\vm \odot\exp\vlambda}{1+\vm^\top\exp\vlambda} \quad \in[0, 1]^d\\
    \mathbb{L}_{\pert}(\pert, \vlambda) &= (1 - \vm^\top\vlambda_\Delta) \norm{\pert} + \vlambda_\Delta^\top \mathcal{L}(\vx+\pert,\ytrue).
\end{aligned}
\end{equation}
By replacing $\vm$ by $\tilde{\vm}^{(t)}$ in each iteration, we obtain the adaptive constraint masking described in \autoref{sec:constraint_strategies}.

The step-size is set to 0.01 for the primal variables and 0.1 for the dual variables with the same exponential and linear decays respectively, as in \cite{matyasko2021pdpgd}. The dual variable is initialized such that the ratio of the weight of the norm term and the constraints terms is $1$. From the above equations, we can derive the initial $\vlambda^{(0)}\in\R^d$ from the ratio $r\in\R_{++}$:
\begin{equation}
\begin{aligned}
    \frac{1 - \vm^\top\vlambda_\Delta^{(0)}}{\vm^\top\vlambda_\Delta^{(0)}}=r
    &\Leftrightarrow 1 - \vm^\top\vlambda_\Delta^{(0)} = r \vm^\top\vlambda_\Delta^{(0)}\\
    &\Leftrightarrow (1 + r) \vm^\top\vlambda_\Delta^{(0)} = 1\\
    &\Leftrightarrow (1 + r)\frac{\vm^\top\exp\vlambda^{(0)}}{1+\vm^\top\exp\vlambda^{(0)}} = 1\\
    &\begin{multlined}\Leftrightarrow (1 + r)\vm^\top\exp\vlambda^{(0)} =\\ 1 + \vm^\top\exp\vlambda^{(0)}\end{multlined}\\
    &\Leftrightarrow \vm^\top\exp\vlambda^{(0)} = \frac{1}{r}
\end{aligned}
\end{equation}
Assuming that $\vlambda^{(0)}=\omega\bm{1}_d$ with $\omega\in\R$, we get:
\begin{equation}
\begin{aligned}
    \vm^\top\exp\vlambda^{(0)} = \frac{1}{r}
    &\Leftrightarrow \norm{\vm}_1\exp \omega = \frac{1}{r}\\
    &\Leftrightarrow \omega = -\log(r\norm{\vm}_1)
\end{aligned}
\end{equation}

\paragraph{Ablation}

To test the effect of these modifications on PDPGD and ALMA $\prox{}$, we perform an ablation study on the masking strategy. We perform this experiment with DeepLabV3+ ResNet-50 on Cityscapes. We compare three different constraint masking strategies:
\begin{itemize}
    \item masking only the unlabeled regions, denoted by $\vm$;
    \item masking the unlabeled regions and $(100 - \nu)\%$ of the largest constraints, denoted by $\tilde{\vm}$;
    \item masking the unlabeled regions and linearly decreasing the fraction of constraints to reach $\nu\%$ at the last iteration, corresponding to strategy described in Equation \eqref{eq:constraint_mask}, denoted by $\tilde{\vm}^{(t)}$.
\end{itemize}

\begin{figure}
    \centering
    \includegraphics[width=\columnwidth]{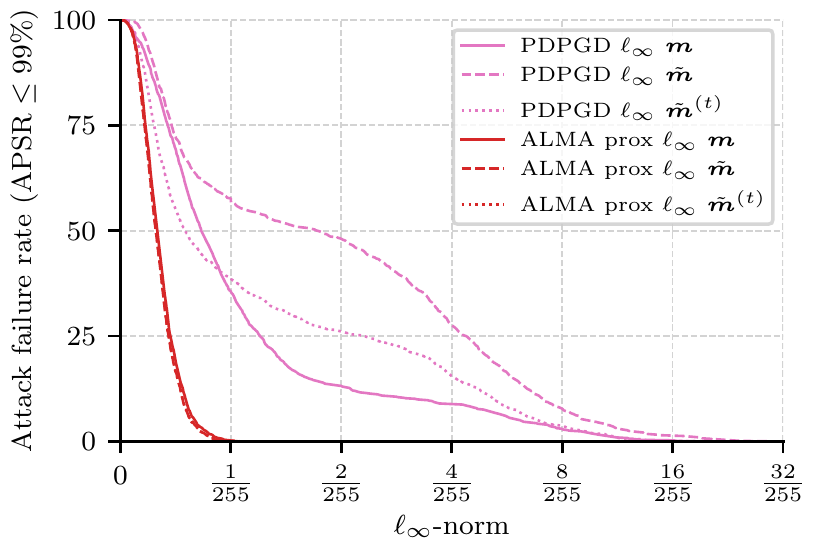}
    \caption{Influence of the constraint masking strategies on PDPGD and ALMA $\prox{}$ for untargeted attacks on Cityscapes with DeepLabV3+ ResNet-50. $\vm$ corresponds to masking the pixels with no labels, $\tilde{\vm}$ corresponds to additionally masking the top $(100 - \nu)\%$ constraints, and $\tilde{\vm}^{(t)}$ corresponds to the strategy in Equation~\eqref{eq:constraint_mask}. The curves for PDGD and ALMA $\prox{}$ $\tilde{\vm}$ and $\tilde{\vm}^{(t)}$ overlap.}
    \label{fig:ablation_constraint_masking}
\end{figure}

The results of this experiment are provided in \autoref{fig:ablation_constraint_masking}.
While the effect is small for this particular model and dataset, the $\tilde{vm}$ and $\tilde{vm}^{(t)}$ strategies do result in smaller perturbations. However, PDPGD does not seem to benefit from the more advanced masking strategies. It obtains the best results when masking only the unlabeled regions. This issue comes from the projection of the dual variables onto the $d$-simplex $\eqref{eq:pdpgd_loss}$: as different constraints get discarded in subsequent iterations, their relative weights vary drastically, leading to oscillations of the dual variables. This phenomenon does not occur with an Augmented Lagrangian based attacks, as the penalty multipliers are not projected together. This does not create a dependency between the multipliers, resulting in a more stable optimization.
Therefore, for the experiments, ALMA $\prox{}$ is used with the adaptive masking strategy with a linear decay, whereas PDPGD is used with the unlabeled region masking only.

\newpage
\section{Attack complexities and run-times}
\label{sec:attack_complexities}

\Autoref{tab:complexities, tab:runtimes} report the average complexities (in terms of forward and backward) and run-time per sample for the attacks. The run-times for the targeted attacks are similar to their targeted variant, as the only difference lies in the loss, which is far from being the most computationally expensive part of the attacks. These results show that ALMA $\prox{}$ has slightly higher run-times compared to the other attacks with similar budgets, while being more effective.
Note that some variance come from the fact that the experiments were run on a shared compute cluster.

\begin{table}[h]
    \centering
    \scriptsize
    \begin{tabular}{lrr}
        Attack & Forwards & Backwards\\
        \midrule
        I-FGSM \cite{kurakin2017adversarial} $13\times 20$ and MI-FGSM \cite{dong2018boosting} & 260 & 260\\
        PGD $13\times 40$ and $13\times 4\times 10$ \cite{madry2018towards} & 520 & 520\\
        DAG $\eta=0.003$ \cite{xie2017adversarial} & 63 & 62\\
        DAG $\eta=0.001$ \cite{xie2017adversarial} & 155 & 154\\
        FMN $\linf$ \cite{pintor2021fast} & 500 & 500\\
        PDPGD $\linf$ \cite{matyasko2021pdpgd} & 500 & 500\\
        ALMA $\mathrm{prox}$ & 500 & 500\\
        \bottomrule
    \end{tabular}
    \caption{Average complexity of the attacks in terms of number of forward and backward propagations. All the attacks, except I-FGSM, MI-FGSM and PGD have a 500 iteration budget. DAG is the only attack that uses an early stopping criterion.}
    \label{tab:complexities}
\end{table}

\begin{table*}[h]
    \footnotesize
    \centering
    \resizebox{\textwidth}{!}{
    \begin{tabular}{llrrrrcrrrrr}
        & & \multicolumn{4}{c}{Pascal VOC 2012} & &\multicolumn{5}{c}{Cityscapes} \\
        \cmidrule{3-6}\cmidrule{8-12}
        & Attack & \makecell{DeepLabV3+\\ResNet-50} & \makecell{DeepLabV3+\\ResNet-101} & \makecell{FCN\\ HRNetV2 W48} & \makecell{DeepLabV3\\DDC-AT} & & \makecell{DeepLabV3+\\ResNet-50} & \makecell{FCN\\HRNetV2 W48} & \makecell{SegFormer\\MiT-B0} & \makecell{SegFormer\\MiT-B3} & \makecell{DeepLabV3\\DDC-AT}\\
        \midrule
        \multirow{11}{*}{\begin{turn}{90}Untargeted\end{turn}}
        & I-FGSM $13\times 20$ \cite{kurakin2017adversarial}          & 13.3 & 19.2 & 28.8 &   -- &  & 87.0  & 43.2 & 31.5 & 83.4  &  --  \\
        & MI-FGSM $13\times 20$ \cite{dong2018boosting}               & 13.3 & 19.2 & 29.1 &   -- &  & 87.6  & 41.5 & 31.1 & 83.2  &  --  \\
        & PGD CE $13\times 40$ \cite{madry2018towards}                & 26.4 & 38.2 & 57.0 &   -- &  & 171.9 & 81.9 & 61.7 & 164.6 &  --  \\
        & PGD CE $13\times 4\times 10$ \cite{madry2018towards}        & 26.6 & 38.3 & 57.2 &   -- &  & 174.1 & 81.9 & 61.6 & 164.5 &  --  \\
        & PGD DLR $13\times 40$ \cite{madry2018towards}               & 27.7 & 39.3 & 60.9 &   -- &  & 179.5 & 88.1 & 67.6 & 170.7 &  --  \\
        & PGD DLR $13\times 4\times 10$ \cite{madry2018towards}       & 27.9 & 39.4 & 57.6 &   -- &  & 179.3 & 88.2 & 67.6 & 170.6 &  --  \\
        & DAG $\eta=0.003$ \cite{xie2017adversarial}                  & 2.1  & 5.2  & 6.0  & 7.0  &  & 19.5  & 14.4 & 9.2  & 49.2  & 22.4 \\
        & DAG $\eta=0.001$ \cite{xie2017adversarial}                  & 4.8  & 12.9 & 14.8 & 12.7 &  & 45.5  & 37.3 & 23.4 & 107.3 & 44.7 \\
        & FMN $\linf$ \cite{pintor2021fast}                           & 24.4 & 32.0 & 41.3 & 24   &  & 155.2 & 79.1 & 59.1 & 158.2 & 64.5 \\
        & PDPGD $\linf$ \cite{matyasko2021pdpgd}                      & 28.1 & 35.5 & 41.0 & 25.5 &  & 161.0 & 84.5 & 64.1 & 162.4 & 64.7 \\
        & ALMA $\mathrm{prox}$                                        & 28.1 & 36.1 & 43.4 & 28   &  & 161.9 & 96.8 & 77.6 & 176.3 & 69.8 \\
        \midrule
        \multirow{11}{*}{\begin{turn}{90}Targeted\end{turn}}
        & I-FGSM $13\times 20$ \cite{kurakin2017adversarial}          & 13.3 & 19.2 & 29.8 &   -- &  & 87.5  & 41.4 & 31.1 & 83.0  &   -- \\
        & MI-FGSM $13\times 20$ \cite{dong2018boosting}               & 13.3 & 19.4 & 28.8 &   -- &  & 87.8  & 41.4 & 31.1 & 83.1  &   -- \\
        & PGD CE $13\times 40$ \cite{madry2018towards}                & 27.0 & 38.1 & 56.8 &   -- &  & 171.9 & 81.9 & 61.4 & 165.0 &   -- \\
        & PGD CE $13\times 4\times 10$ \cite{madry2018towards}        & 26.7 & 38.4 & 56.5 &   -- &  & 171.9 & 82.0 & 61.5 & 164.3 &   -- \\
        & PGD DLR $13\times 40$ \cite{madry2018towards}               & 27.7 & 39.4 & 58.9 &   -- &  & 179.6 & 88.2 & 68.4 & 172.0 &   -- \\
        & PGD DLR $13\times 4\times 10$ \cite{madry2018towards}       & 27.8 & 39.8 & 58.7 &   -- &  & 179.4 & 88.2 & 69.7 & 171.0 &   -- \\
        & DAG $\eta=0.003$ \cite{xie2017adversarial}                  & 0.8  & 1.5  & 1.9  & 2.2  &  & 47.6  & 18.6 & 21.9 & 54.1  & 44.5 \\
        & DAG $\eta=0.001$ \cite{xie2017adversarial}                  & 2.1  & 4.2  & 4.6  & 4.8  &  & 111.7 & 44.9 & 52.0 & 132.9 & 61.0 \\
        & FMN $\linf$ \cite{pintor2021fast}                           & 24.4 & 32.1 & 41.1 & 25.4 &  & 155.1 & 79.2 & 58.7 & 157.9 & 60.4 \\
        & PDPGD $\linf$ \cite{matyasko2021pdpgd}                      & 27.6 & 36.9 & 39.0 & 26.8 &  & 158.8 & 81.4 & 64.1 & 159.1 & 63.2 \\
        & ALMA $\mathrm{prox}$                                        & 28.4 & 36.3 & 43.0 & 28.5 &  & 161.9 & 97.2 & 78.4 & 179.5 & 67.6 \\
        \bottomrule
    \end{tabular}
    }
    \caption{Average run-times per image for the attacks, in seconds.}
    \label{tab:runtimes}
\end{table*}

\section{Complete attack results}
\label{sec:attack_results}

\begin{figure*}
    \centering
    \includegraphics[width=0.9\columnwidth]{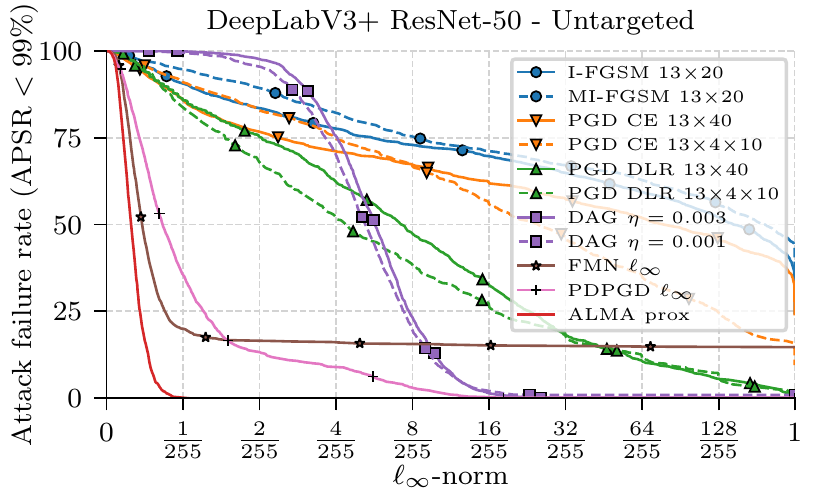}
    \hspace{1em}
    \includegraphics[width=0.9\columnwidth]{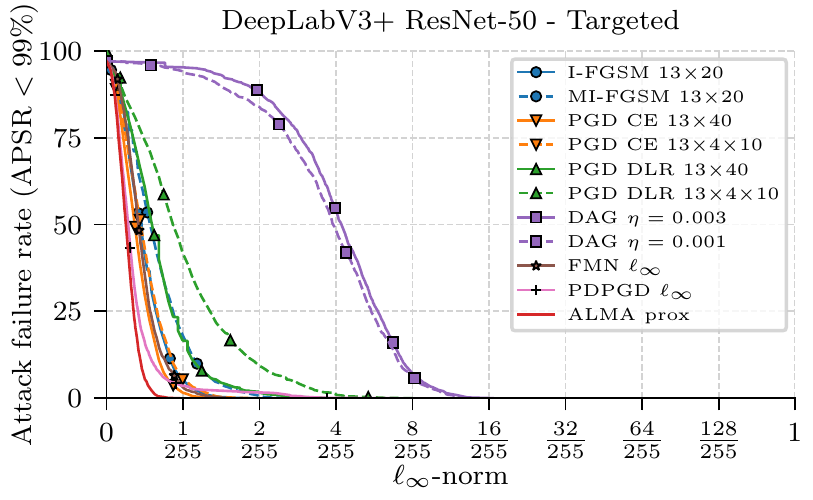}\\
    \includegraphics[width=0.9\columnwidth]{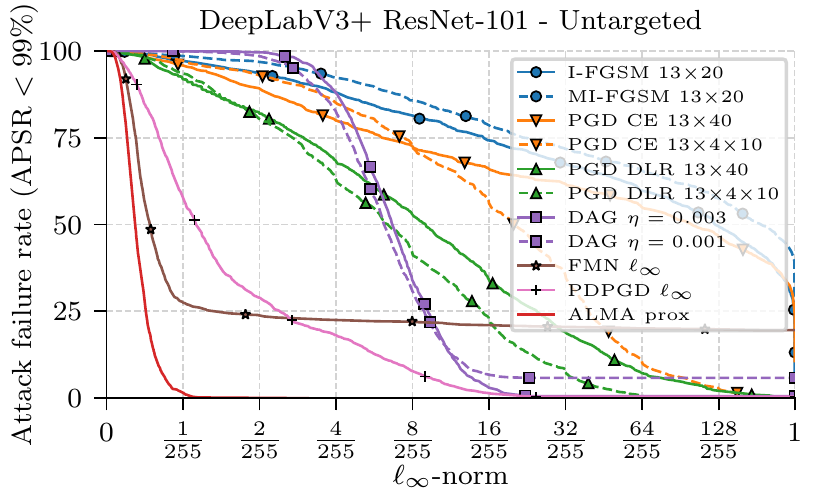}
    \hspace{1em}
    \includegraphics[width=0.9\columnwidth]{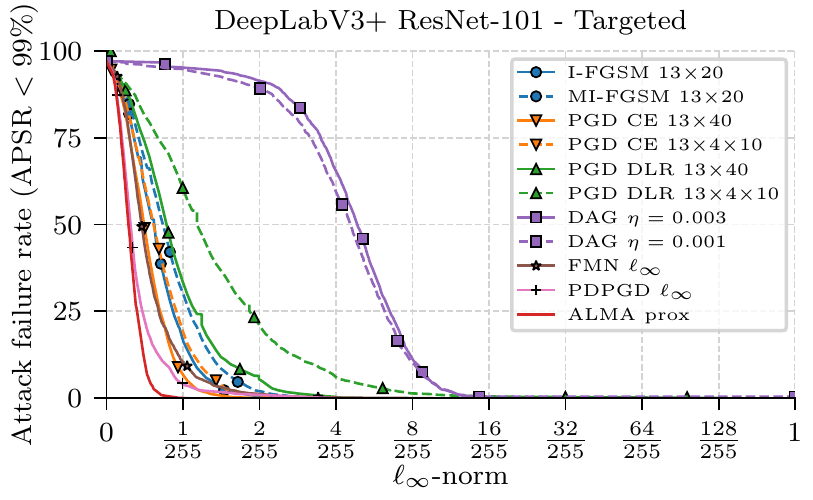}\\
    \includegraphics[width=0.9\columnwidth]{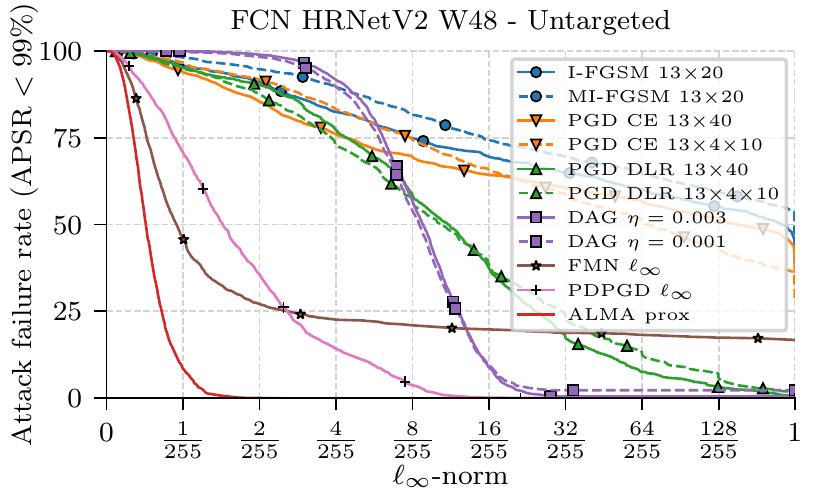}
    \hspace{1em}
    \includegraphics[width=0.9\columnwidth]{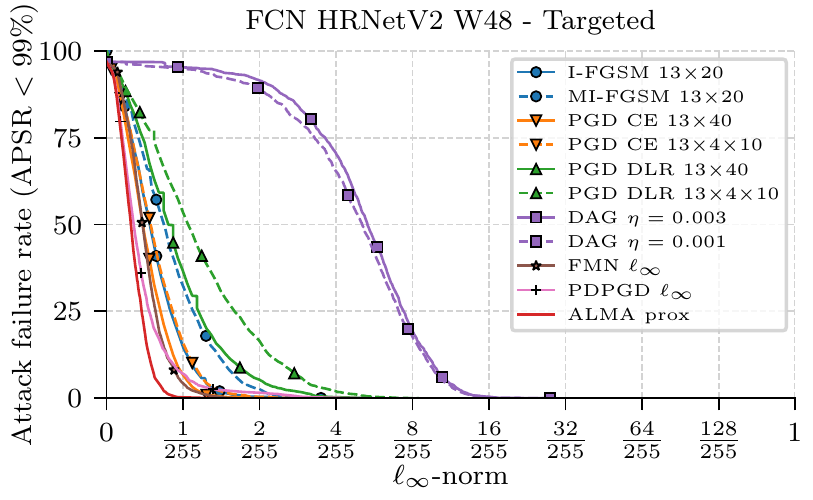}\\
    \includegraphics[width=0.9\columnwidth]{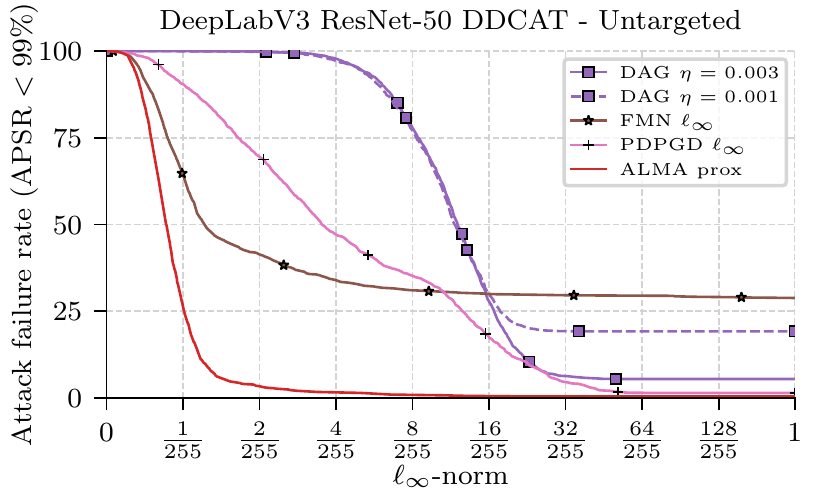}
    \hspace{1em}
    \includegraphics[width=0.9\columnwidth]{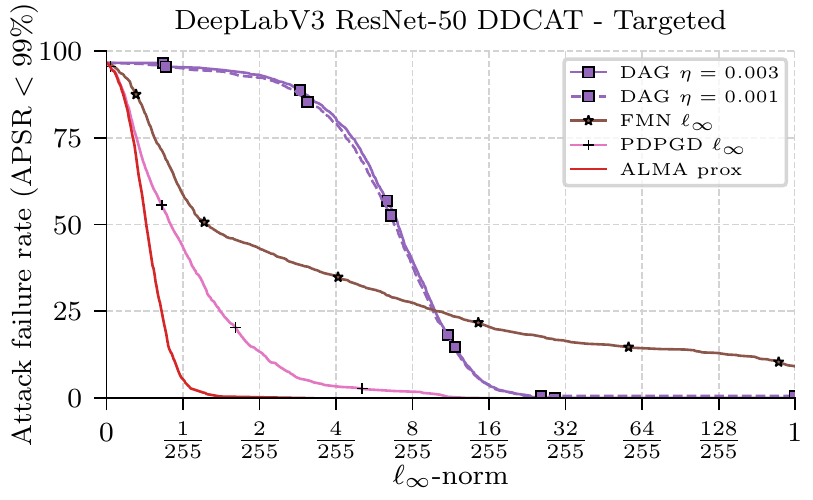}\\
    \caption{Percentage of unsuccessful $\linf$ attacks on \textbf{Pascal VOC 2012} (\ie with $\mathrm{APSR} \leq 99\%$). A stronger attack has a lower curve; a more robust model has a higher curve. Horizontal axis is linear on $[0, \nicefrac{2}{255}]$ and logarithmic on $[\nicefrac{2}{255}, 1]$.}
    \label{fig:linf_pascal}
\end{figure*}

\begin{figure*}
    \centering
    \includegraphics[width=0.85\columnwidth]{figures/cityscapes_deeplabv3plus_resnet50_linf_untargeted.pdf}
    \hspace{1em}
    \includegraphics[width=0.85\columnwidth]{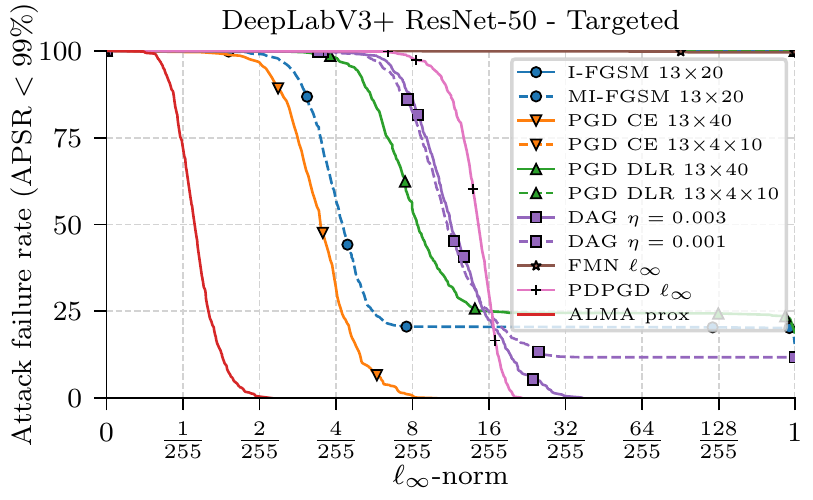}\\
    \includegraphics[width=0.85\columnwidth]{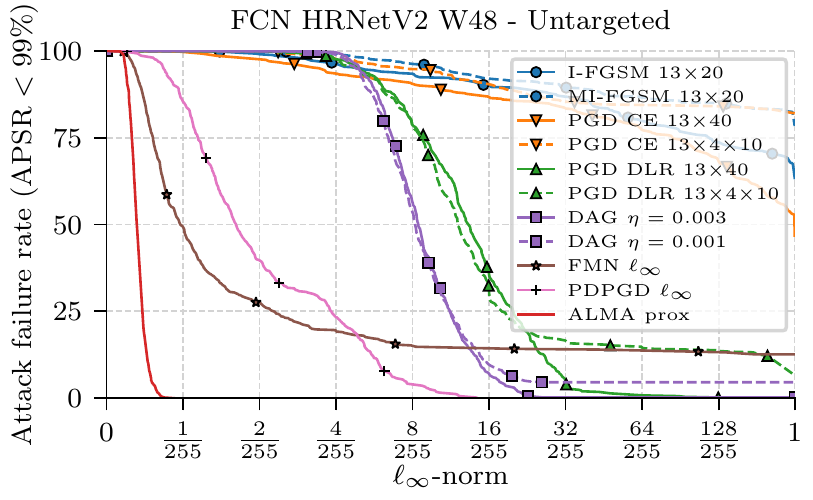}
    \hspace{1em}
    \includegraphics[width=0.85\columnwidth]{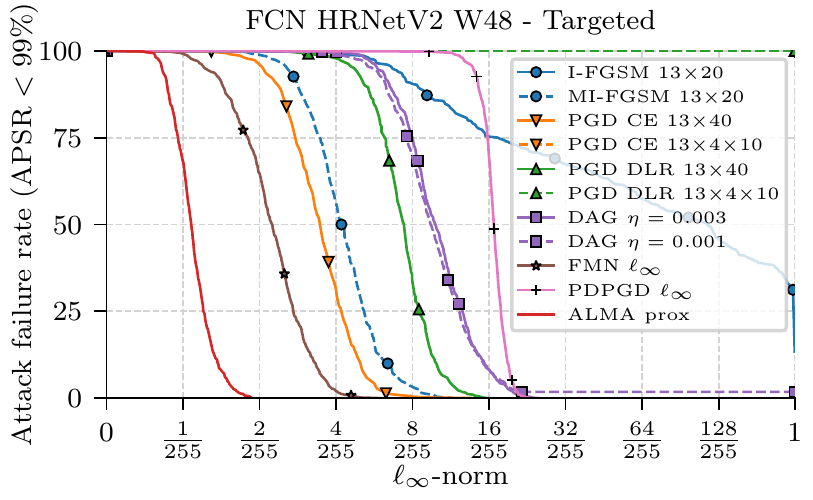}\\
    \includegraphics[width=0.85\columnwidth]{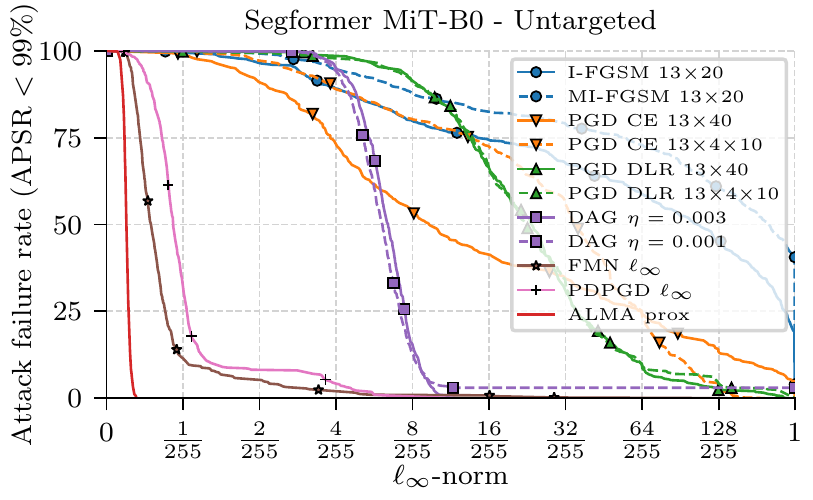}
    \hspace{1em}
    \includegraphics[width=0.85\columnwidth]{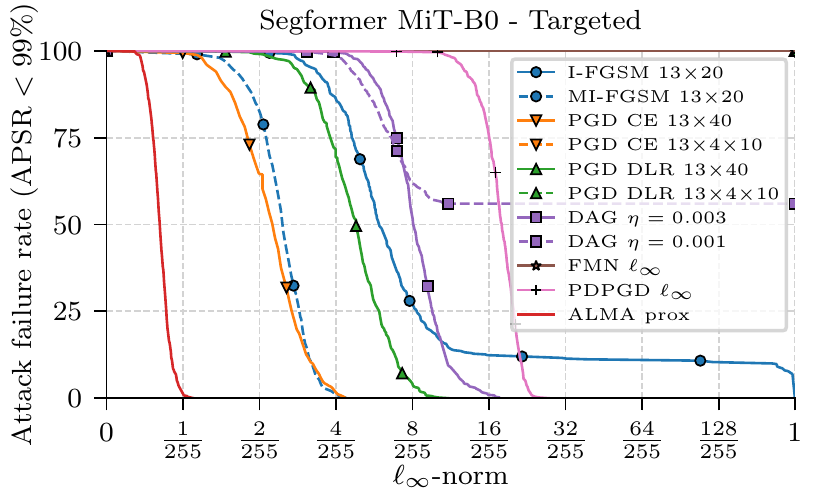}\\
    \includegraphics[width=0.85\columnwidth]{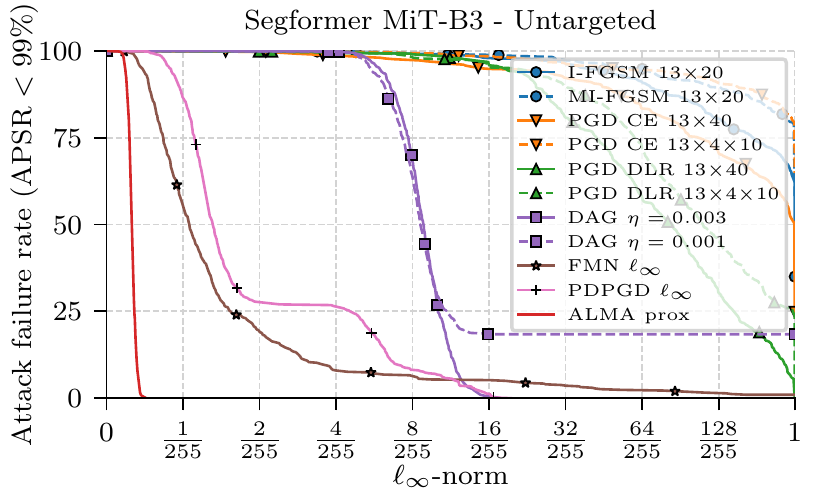}
    \hspace{1em}
    \includegraphics[width=0.85\columnwidth]{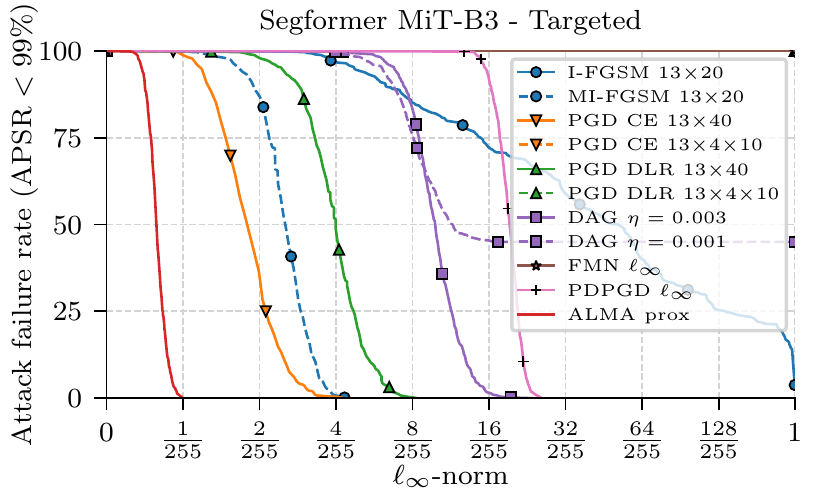}\\
    \includegraphics[width=0.85\columnwidth]{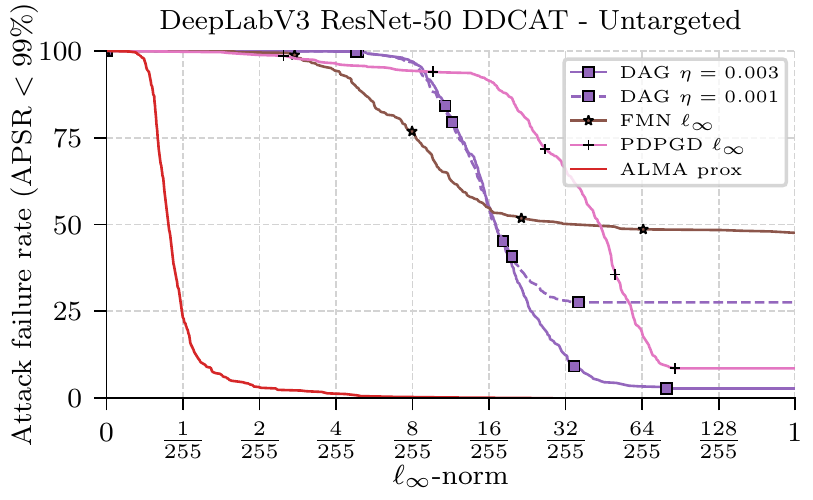}
    \hspace{1em}
    \includegraphics[width=0.85\columnwidth]{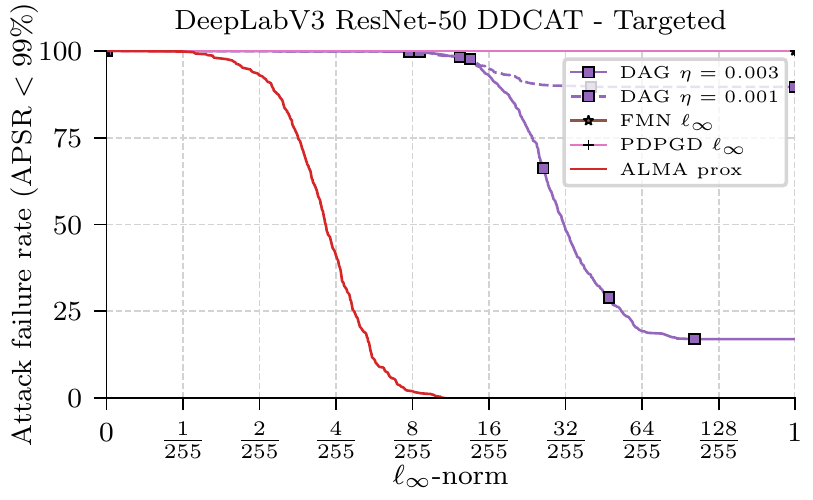}\\
    \caption{Percentage of unsuccessful $\linf$ attacks on \textbf{Cityscapes} (\ie with $\mathrm{APSR} \leq 99\%$). A stronger attack has a lower curve; a more robust model has a higher curve. Horizontal axis is linear on $[0, \nicefrac{2}{255}]$ and logarithmic on $[\nicefrac{2}{255}, 1]$.}
    \label{fig:linf_cityscapes}
\end{figure*}

\autoref{tab:median_average_pert} reports the median and average $\linf$-norm (multiplied by 255 for readability) of the perturbations produced by the attacks for all regular models. As in \autoref{sec:results}, the perturbation norm is considered to be 1 for unsuccessful attacks. This means that attack with less than 50\% success have a 1 (\ie 255) median $\linf$-norm in the table.
Additionally, \Autoref{fig:linf_pascal, fig:linf_cityscapes} show the percentage of unsuccessful attacks on Pascal VOC 2012 and Cityscapes for all models considered.

\begin{table*}
    \footnotesize
    \centering
    \resizebox{\textwidth}{!}{
    \begin{tabular}{ll*{6}{r}c*{8}{r}}
        & & \multicolumn{6}{c}{Pascal VOC 2012} & & \multicolumn{8}{c}{Cityscapes} \\
        \cmidrule{3-8}\cmidrule{10-17}
        & Attack & \multicolumn{2}{c}{\makecell{DeepLabV3+\\ResNet-50}} &  \multicolumn{2}{c}{\makecell{DeepLabV3+\\ResNet-101}} &  \multicolumn{2}{c}{\makecell{FCN\\ HRNetV2 W48}} & &  \multicolumn{2}{c}{\makecell{DeepLabV3+\\ResNet-50}} & \multicolumn{2}{c}{\makecell{FCN\\HRNetV2 W48}} &  \multicolumn{2}{c}{\makecell{SegFormer\\MiT-B0}} &  \multicolumn{2}{c}{\makecell{SegFormer\\MiT-B3}} \\
        \midrule
        \multirow{11}{*}{\begin{turn}{90}Untargeted\end{turn}}
        & I-FGSM $13\times 20$ \cite{kurakin2017adversarial}        & 146.32 & 136.55 & 124.14 & 131.34 & 227.11 & 145.69 &  & 255.00 & 242.15 & 255.00 & 194.98 & 106.34 & 121.12 & 255.00 & 208.97 \\
        & MI-FGSM $13\times 20$ \cite{dong2018boosting}             & 195.35 & 145.96 & 188.37 & 150.13 & 255.00 & 157.63 &  & 255.00 & 244.43 & 255.00 & 218.91 & 202.15 & 159.52 & 255.00 & 228.43 \\
        & PGD CE $13\times 40$ \cite{madry2018towards}              & 80.10 & 120.93 & 100.42 & 123.17 & 152.08 & 136.05 &  & 255.00 & 236.77 & 255.00 & 176.94 & 9.31 & 48.10 & 255.00 & 188.15 \\
        & PGD CE $13\times 4\times 10$ \cite{madry2018towards}      & 24.90 & 74.15 & 20.05 & 30.28 & 66.93 & 118.76 &  & 255.00 & 244.92 & 255.00 & 216.38 & 34.77 & 42.49 & 255.00 & 231.51 \\
        & PGD DLR $13\times 40$ \cite{madry2018towards}             & 7.22 & 26.42 & 9.00 & 21.29 & 11.11 & 23.85 &  & 17.75 & 24.23 & 13.29 & 16.15 & 23.53 & 32.34 & 84.22 & 102.80 \\
        & PGD DLR $13\times 4\times 10$ \cite{madry2018towards}     & 4.42 & 24.99 & 6.41 & 11.02 & 10.46 & 29.75 &  & 255.00 & 227.89 & 12.08 & 45.17 & 22.37 & 35.49 & 110.82 & 134.41 \\
        & DAG $\eta=0.003$ \cite{xie2017adversarial}                & 5.69 & 6.63 & 6.65 & 8.74 & 8.80 & 10.61 &  & 6.30 & 7.51 & 8.42 & 9.91 & 6.41 & 6.48 & 8.83 & 9.02 \\
        & DAG $\eta=0.001$ \cite{xie2017adversarial}                & 5.23 & 8.22 & 6.17 & 21.14 & 8.49 & 14.61 &  & 5.95 & 9.39 & 8.20 & 20.29 & 6.08 & 13.57 & 8.59 & 53.65 \\
        & FMN $\linf$ \cite{pintor2021fast}                         & 0.46 & 38.34 & 0.56 & 51.60 & 0.91 & 46.39 &  & 1.97 & 96.57 & 0.97 & 35.12 & 0.58 & 1.16 & 1.08 & 6.42 \\
        & PDPGD $\linf$ \cite{matyasko2021pdpgd}                    & 0.73 & 1.77 & 1.17 & 3.49 & 1.52 & 2.43 &  & 1.06 & 2.82 & 1.69 & 2.75 & 0.87 & 1.17 & 1.39 & 3.05 \\
        & ALMA $\mathrm{prox}$                                      & \textbf{0.32} & \textbf{0.34} & \textbf{0.37} & \textbf{0.41} & \textbf{0.51} & \textbf{0.56} & & \textbf{0.24} & \textbf{0.26} & \textbf{0.40} & \textbf{0.41} & \textbf{0.26} & \textbf{0.26} & \textbf{0.33} & \textbf{0.33} \\
        \midrule
        \multirow{11}{*}{\begin{turn}{90}Targeted\end{turn}}
        & I-FGSM $13\times 20$ \cite{kurakin2017adversarial}        & 0.47 & 0.50 & 0.65 & 0.67 & 0.59 & 0.64 &  & 255.00 & 255.00 & 115.76 & 128.65 & 5.91 & 31.96 & 51.73 & 88.24 \\
        & MI-FGSM $13\times 20$ \cite{dong2018boosting}             & 0.59 & 0.66 & 0.77 & 0.82 & 0.77 & 0.85 &  & 4.26 & 55.35 & 4.23 & 4.47 & 2.48 & 2.53 & 2.54 & 2.60 \\
        & PGD CE $13\times 40$ \cite{madry2018towards}              & 0.37 & 0.43 & 0.50 & 0.55 & 0.50 & 0.54 &  & 3.49 & 3.71 & 3.49 & 3.62 & 2.30 & 2.34 & 1.87 & 1.92 \\
        & PGD CE $13\times 4\times 10$ \cite{madry2018towards}      & 0.50 & 0.51 & 0.62 & 0.70 & 0.62 & 0.65 &  & 255.00 & 255.00 & 255.00 & 255.00 & 255.00 & 255.00 & 255.00 & 255.00 \\
        & PGD DLR $13\times 40$ \cite{madry2018towards}             & 0.62 & 0.68 & 0.81 & 0.92 & 0.81 & 0.94 &  & 8.37 & 67.86 & 7.41 & 7.52 & 4.79 & 4.96 & 3.98 & 4.09 \\
        & PGD DLR $13\times 4\times 10$ \cite{madry2018towards}     & 0.87 & 1.07 & 1.18 & 2.03 & 1.12 & 1.28 &  & 255.00 & 255.00 & 255.00 & 255.00 & 255.00 & 255.00 & 255.00 & 255.00 \\
        & DAG $\eta=0.003$ \cite{xie2017adversarial}                & 4.21 & 4.50 & 4.84 & 5.09 & 5.32 & 5.66 &  & 11.34 & 12.96 & 9.87 & 10.49 & 8.05 & 8.44 & 9.82 & 10.06 \\
        & DAG $\eta=0.001$ \cite{xie2017adversarial}                & 3.92 & 4.21 & 4.55 & 5.80 & 5.07 & 5.36 &  & 10.96 & 40.28 & 9.43 & 14.42 & 255.00 & 145.61 & 11.53 & 119.47 \\
        & FMN $\linf$ \cite{pintor2021fast}                         & 0.42 & 0.45 & 0.46 & 0.56 & 0.47 & 0.49 &  & 255.00 & 254.36 & 2.25 & 2.34 & 255.00 & 255.00 & 255.00 & 255.00 \\
        & PDPGD $\linf$ \cite{matyasko2021pdpgd}                    & 0.28 & 0.36 & 0.31 & 0.40 & 0.35 & 0.46 &  & 14.51 & 14.41 & 16.71 & 16.75 & 17.93 & 17.87 & 19.26 & 19.20 \\
        & ALMA $\mathrm{prox}$                                      & \textbf{0.25} & \textbf{0.26} & \textbf{0.29} & \textbf{0.30} & \textbf{0.32} & \textbf{0.34} & & \textbf{1.15} & \textbf{1.17} & \textbf{1.11} & \textbf{1.12} & \textbf{0.70} & \textbf{0.70} & \textbf{0.65} & \textbf{0.66} \\
        \bottomrule
    \end{tabular}
    }
    \caption{Median and average $\norm{\pert}_\infty{\times}255$ for each adversarial attack on Pascal VOC 2012 and Cityscapes for the regular models.}
    \label{tab:median_average_pert}
\end{table*}

